\documentclass{article}

\PassOptionsToPackage{numbers}{natbib}
\usepackage[preprint]{neurips_2024}
\usepackage{subcaption}
\usepackage{graphicx}
\usepackage{float}
\usepackage[justification=centering]{caption}

\usepackage[utf8]{inputenc} 

\usepackage[T1]{fontenc}    
\usepackage{hyperref}       
\usepackage{url}            
\usepackage{booktabs}       
\usepackage{amsfonts}       
\usepackage{nicefrac}       
\usepackage{microtype}      
\usepackage{xcolor}         

\title{Efficiently Dispatching Flash Attention For Partially Filled Attention Masks}

\author{
Agniv Sharma$^{1, 2}$\quad Jonas Geiping$^{2, 3, 4}$ \vspace{8pt}\\
$^{1}$University of T\"ubingen\qquad $^{2}$ T\"ubingen AI Center\qquad \\
$^{3}$ ELLIS Institute T\"ubingen \qquad \\
$^{4}$Max Planck Institute for Intelligent Systems
}

\begin{document}

\maketitle

\begin{abstract}
  Transformers are widely used across various applications, many of which yield sparse or partially filled attention matrices. Examples include attention masks designed to reduce the quadratic complexity of attention, sequence packing techniques, and recent innovations like tree masking for fast validation in MEDUSA. Despite the inherent sparsity in these matrices, the state-of-the-art algorithm Flash Attention still processes them with quadratic complexity as though they were dense. In this paper, we introduce \textbf{Binary Block Masking}, a highly efficient modification that enhances Flash Attention by making it mask-aware. We further propose two optimizations: one tailored for masks with contiguous non-zero patterns and another for extremely sparse masks. Our experiments on attention masks derived from real-world scenarios demonstrate up to a 9x runtime improvement. The implementation will be publicly released to foster further research and application.
\end{abstract}

\section{Introduction}
\label{sec:Intro}

\begin{figure}
    \centering
    \begin{subfigure}[b]{0.32\textwidth}
        \centering
        \includegraphics[width=\textwidth]{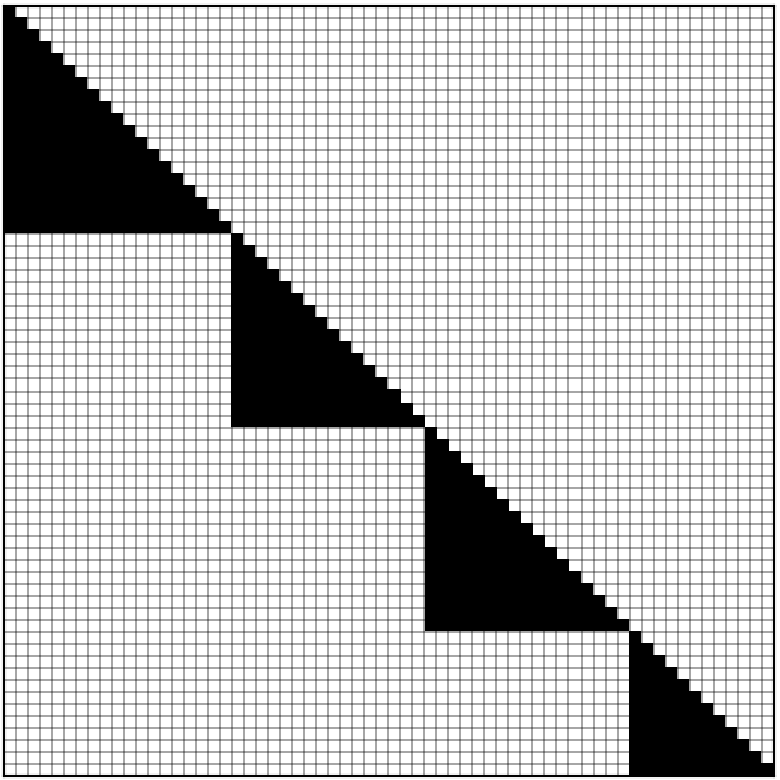}
        \caption{Original Attention Mask}
    \end{subfigure}
    \hfill
    \begin{subfigure}[b]{0.32\textwidth}
        \centering
        \includegraphics[width=\textwidth]{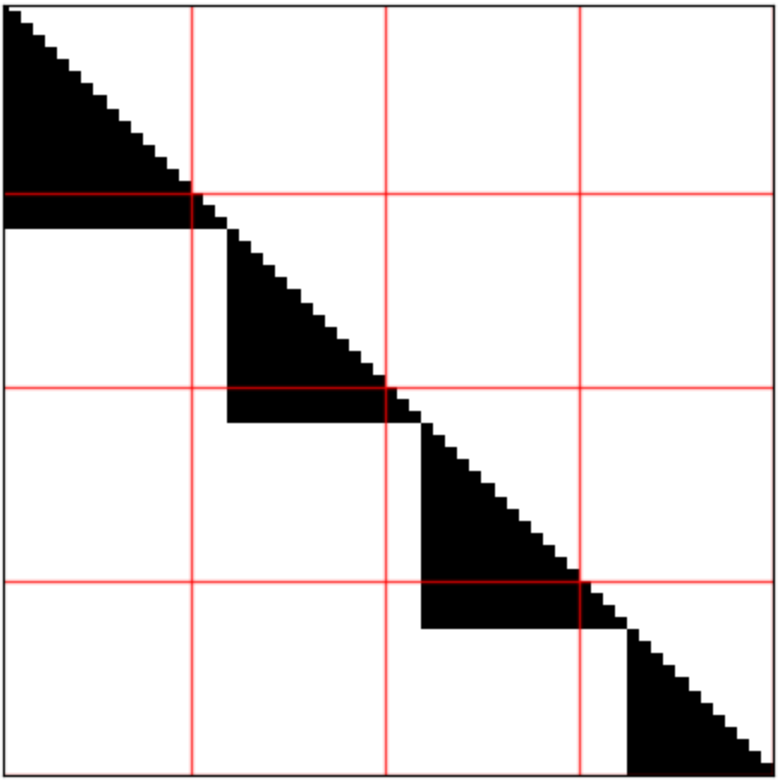}
        \caption{Mask with Binary Blocks}
    \end{subfigure}
    \hfill
    \begin{subfigure}[b]{0.32\textwidth}
        \centering
        \includegraphics[width=\textwidth]{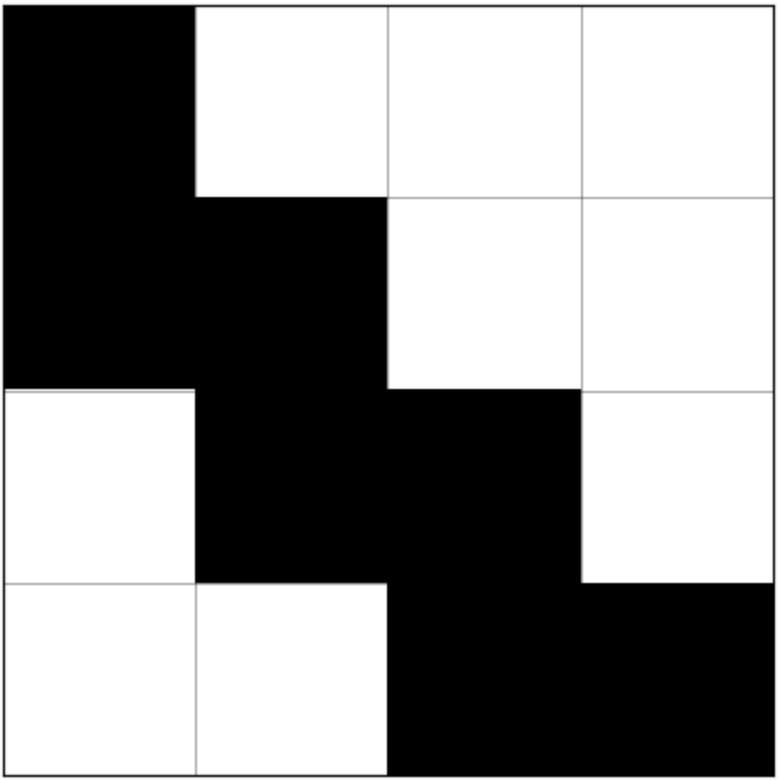}
        \caption{Binary Block Matrix}
    \end{subfigure}
    \hfill
    \begin{subfigure}[b]{\textwidth}
        \centering
        \includegraphics[width=0.75\textwidth]{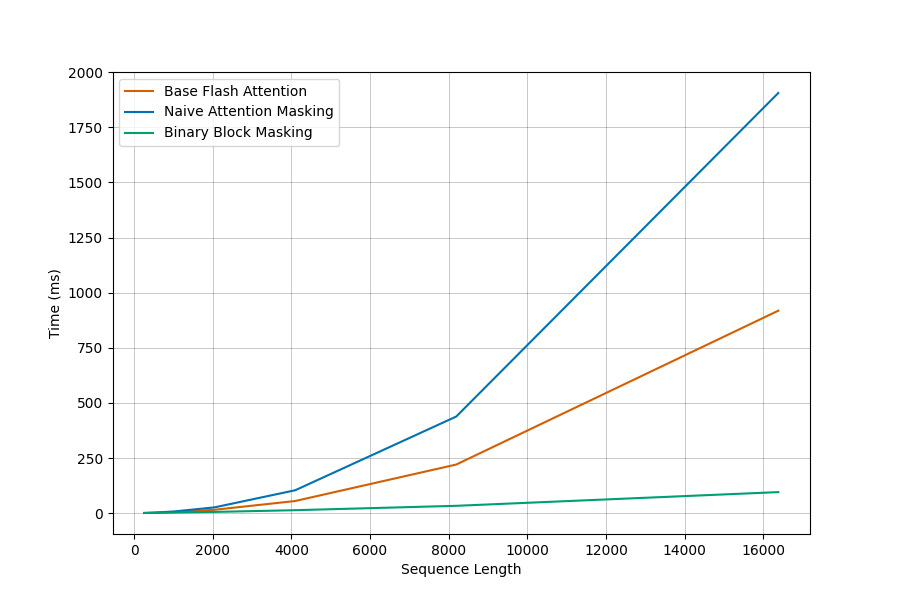}
        \caption{Performance Comparison}
    \end{subfigure}
    
    \caption{Proposed algorithm for masking flash attention}
    \label{fig:proposed_algorithm}
\end{figure}

Transformers \cite{Transformers} revolutionized sequence modeling by using self-attention mechanisms, enabling efficient parallelization and handling of long-range dependencies. A key component of transformers is the attention mechanism. However, in many practical scenarios—such as when processing very long sequences, reducing computational costs, or working with structured data like graphs — it becomes advantageous to restrict the range of elements each item can attend to. This leads to the development of masked attention mechanisms, which limit interactions to a subset of elements.

One application that leads to such partially filled masks is the creation of efficient transformers. Approaches like \textit{LongFormer} and \textit{BigBird} \cite{Beltagy2020Longformer, zaheer2020bigbird} employ fixed sparsity patterns to alleviate the quadratic complexity inherent in self-attention, while other approaches exploit low-rank approximations to achieve similar objectives \cite{wang2020linformer, dao2019butterfly}. 
Despite the theoretically lower computational complexity of these methods, many results fail to translate to reduced wall-clock runtimes.
The currently prevalent practical solution is instead to simply implement dense attention more efficiently. Flash Attention  \cite{NEURIPS2022_FlashAttn1, dao2023flashattention2, shah2024flashattention3} is a such a hardware-aware algorithm sufficient to boost the speed of dense attention and obviate the need for a more efficient attention in many applications.

However, because the initial flash attention implementation was written with only causal masks in mind, research using alternative attention patterns has slowed down.
A previous approach to remedy this, \cite{pagliardini2023fasterCausal}, enables a few types of mask in flash attention, but only considers a small subset of cases and requires the user to carefully change the GPU kernel parameters for every new mask. To overcome these limitations, we introduce \textbf{Binary Block Masking} (\texttt{BinBlkMsk}) - a novel algorithm that extends Flash Attention to support any attention masks while also being user-friendly. Our approach builds on two key insights: (1) Flash Attention processes attention matrices in blocks, so we only need to handle blocks containing at least one non-zero mask value, and (2) these blocks are independent and can be preprocessed in parallel with minimal runtime overhead and then shared across multiple heads, layers, or runs depending on the use case. We further enhance the algorithm's efficiency for masks with contiguous non-zero blocks. Finally, we discuss how off-the-shelf graph optimization algorithms allow these gains to be realized even for relatively unstructured attention patterns, which can be preprocessed into low bandwidth layouts to improve block sparsity.

Our method introduces efficient support for custom masks in flash attention. With this improvement, we achieve up to a 9x reduction in runtime for real-world sparse and partially filled attention matrices. We validate the performance gains of our method using masks derived from three practical applications: packed fine-tuning on the ALPACA dataset \cite{alpaca}, tree attention masks for MEDUSA \cite{cai2024medusa}, and sparse attention matrices from Longformer\cite{Beltagy2020Longformer}.

Our method’s key advantages can be summarized as:

\begin{enumerate}

\item \textbf{Universal Mask Compatibility}: Our algorithm is the first to support any custom mask, providing a generalized solution without the need for user adjustments.

\item \textbf{Performance Consistency}: Our implementation significantly outperforms Flash Attention for partially filled or sparse attention masks, while matching its performance for almost filled attention masks.

\end{enumerate}

We implement our algorithm using Triton \cite{tillet2019triton}, leveraging its device agnosticism, readability, and ease of integration. Our implementation and scripts for processing various attention masks will be made available to support further research.

\section{Related Work}
Since their inception, transformers have employed masked attention for various applications. The original transformer\cite{Transformers} uses causal masking to maintain the sequential nature of text, and standard techniques, such as 
in \cite{raffel2020packing}, utilize attention masks to 'pack' multiple sequences into one block during pretraining.
Masking has also been utilized to create efficient transformers \cite{Survey_Transformers, Efficient_survey}. Models like Star Transformer\cite{guo2022startransformer}, Longformer\cite{Beltagy2020Longformer}, Big Bird\cite{zaheer2020bigbird}, and ETC\cite{ainslie-etal-2020-etc} use attention masks with global and local token windows to speed up attention calculations while preserving information flow. Sparse transformers \cite{child2019sparse} use factorized attention mask to attend to different parts of input across different heads and Reformer\cite{Kitaev2020Reformer:} uses s locality-sensitive-hashing (LSH) to select keys which limit attention to keys and queries that collide to the same hash.

However, masked attention is a natural building block in a number of applications across domains, such as graph learning with transformers \cite{buterez2024masked} or selective perception in vision transformers \cite{wei2023sparsifiner}. On the other hand, inference techniques to speed up generation in text generation, such as \cite{cai2024medusa, miao2024specinfer, spector2023tree_batch} all also use some form of a tree attention mask to increase the number of candidates validated at each step of speculative decoding\cite{leviathan2023SpecDecoding1, chen2023SpecDecoding2}. All these approaches utilize masked attention, but ours is the first method to combine the advantages of state-of-the-art Flash Attention with masking, achieving the best of both techniques.

Our method closely relates to two previous approaches: Block Sparse Attention\cite{NEURIPS2022_FlashAttn1} and Faster Causal Attention\cite{pagliardini2023fasterCausal}. In Block Sparse Attention\cite{NEURIPS2022_FlashAttn1}, authors skip processing some blocks randomly, but unlike our approach, they do not use a mask to guide this, resulting in a mask-agnostic method that provides only approximate attention, whereas ours offers exact calculations. Faster Causal Attention\cite{pagliardini2023fasterCausal} handles only two specific masks—QK masks (dropping random query keys) and hash sparse masks—but it lacks flexibility for general masks and requires significant adaptation for new methods. In contrast, our method is generalizable and intuitive across different masking strategies.

\section{Background}
\textbf{Standard Attention} In standard self-attention\cite{Transformers}, the Query (Q) and Key (K) matrices, both of size $N \times D$ (where $N$ is the number of tokens and $D$ is the dimensionality), are multiplied to produce an $N \times N$ attention matrix. This matrix is normalized using softmax to form a probability distribution, which is then multiplied by the Value (V) matrix ($N \times D$) to generate the final output. In multi-head attention, this process is repeated across multiple heads, and typically processed over batches to enhance parallelism and capture diverse patterns.

\textbf{Flash Attention} Flash Attention\cite{NEURIPS2022_FlashAttn1, dao2023flashattention2} improves self-attention efficiency through two key optimizations. First, it computes the attention matrix in blocks using a running softmax, reducing data transfers from High Bandwidth Memory (HBM). Second, it recomputes the attention matrix during the backward pass, saving memory and optimizing IO operations. These techniques make Flash Attention particularly effective for large models and long sequence lengths.

\textbf{Flash Attention with \textit{naive} masking}. Flash Attention currently only supports standard causal masks. Extending Flash Attention to handle custom masks using a naive approach would involve reading the corresponding mask block for each Flash Attention block. While this method would correctly apply the mask, it introduces additional memory accesses for reading the mask and extra computation for applying it, resulting in a slower runtime than the base Flash Attention. We refer to this method as \textit{Naive Attention Masking}, and benchmark its performance to highlight the trade-offs.
\section{Method}
\label{sec:Method}
\textbf{Binary Block Masking} Our method focuses on optimizing Flash Attention by processing only the blocks of the attention matrix that have non-zero entries in their corresponding mask blocks. We first preprocess the attention mask to create a binary matrix, termed the "Binary Block Matrix" (\texttt{BinBlkMat}). This matrix has dimensions $\texttt{N//BLOCKSIZE\_I} \times \texttt{N//BLOCKSIZE\_J} $ where $N$ is the size of the attention mask and $\texttt{BLOCKSIZE\_I}$ and $\texttt{BLOCKSIZE\_J}$ represent the block sizes in the \texttt{row} and \texttt{column} dimensions respectively. Each entry in \texttt{BinBlkMat} is set to 1 if any value in the corresponding mask block is non-zero.

During the attention step, the relevant block is processed only if its corresponding entry in \texttt{BinBlkMat} is non-zero, reducing unnecessary computations. When masks are fixed, \texttt{BinBlkMat} can be precomputed. For dynamically changing masks, \texttt{BinBlkMat} can be calculated in parallel for all blocks and shared across transformer layers and heads. Thus, \texttt{BinBlkMat} introduces minimal runtime overhead while significantly improving the efficiency.

\looseness -1 \textbf{Dense Binary Block Masking} In natural language processing, masks often contain contiguous non-zero blocks. To optimize for such cases, we leverage the insight that only the edges of these contiguous blocks need to be checked, as the center is entirely ones. We introduce two additional arrays, "\texttt{total\_ones}" and "\texttt{offset}", both of size $ \texttt{N//BLOCKSIZE\_I} \times 1$. "\texttt{total\_ones}" stores the number of consecutive blocks filled with ones, while "\texttt{offset}" stores the position of the first such block.

During the attention step, we parallelize across the Q dimension and iterate over blocks of K. Thus, while processing \texttt{i\textsuperscript{th}} block of Q, the mask is read only if the current iteration is less than the \texttt{offset\_i} or greater than the \texttt{offset\_i} + \texttt{total\_ones\_i}, and if the corresponding \texttt{BinBlkMat} value is 1. For iterations within the range of the contiguous block, the mask read is skipped, as it consists entirely of ones. This significantly reduces the number of mask reads from HBM, speeding up processing. Similar to \texttt{BinBlkMat}, the \texttt{total\_ones} and \texttt{offset} arrays can be computed in parallel with minimal overhead and shared across layers and heads.

\begin{figure}
    \centering
    \begin{subfigure}[b]{0.21\textwidth}
        \centering
        \includegraphics[width=\textwidth]{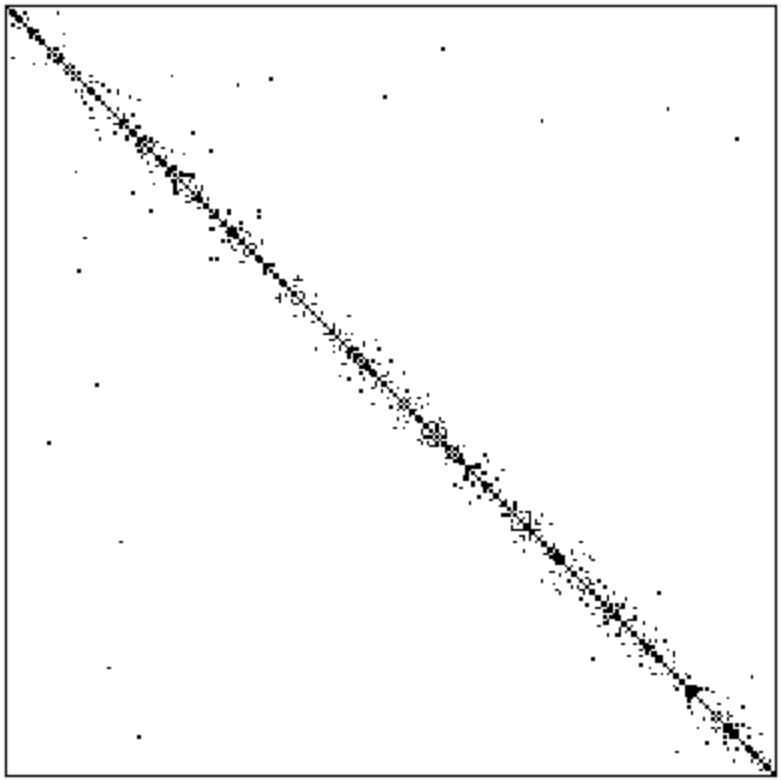}
        \vspace{0.5cm}  
        \includegraphics[width=\textwidth]{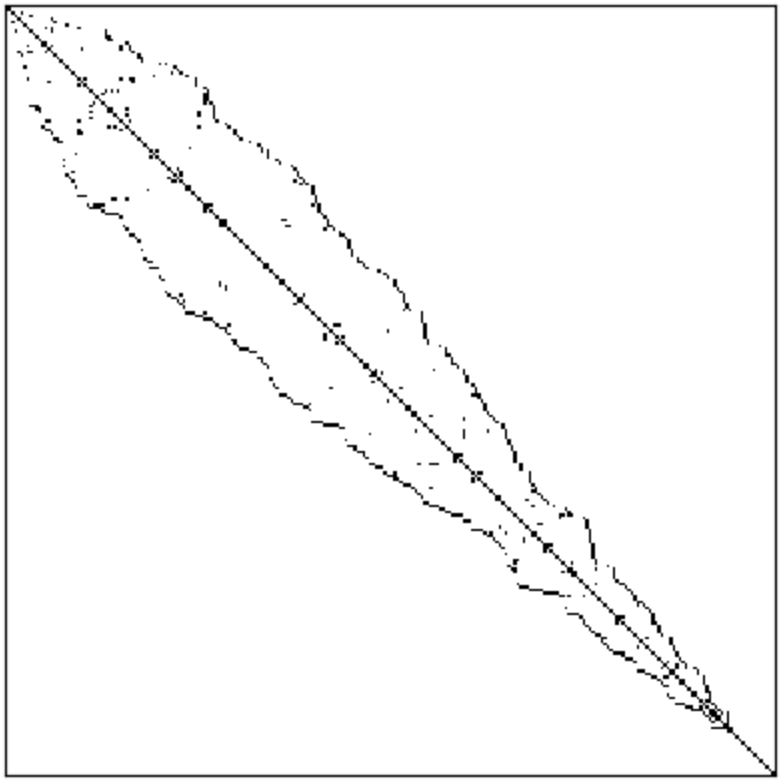}
        \caption{Bandwidth reduction of a sparse mask through RCM}
        \label{fig:RCM_Mask}
    \end{subfigure}
    \hfill
    \begin{subfigure}[b]{0.75\textwidth}
        \centering
        \includegraphics[width=\textwidth]{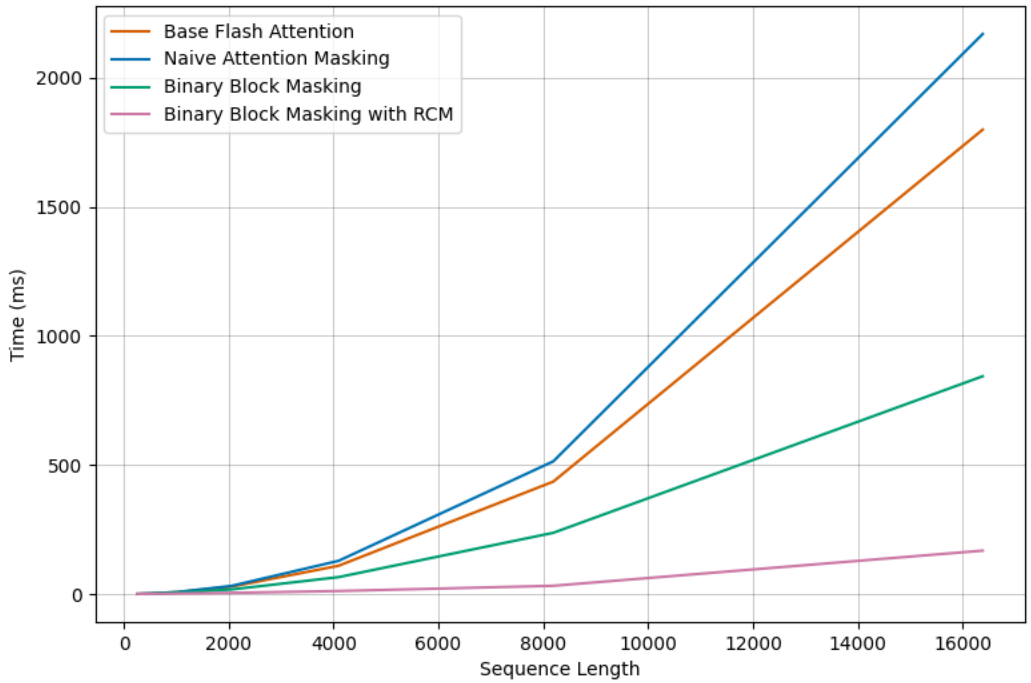}
        \caption{Performance Gains achieved through RCM preprocessing}
        \label{fig:RCM_perf}
    \end{subfigure}
    
    \caption{The result and performance of the RCM bandwidth reduction algorithm when computing a sparse attention mask.}
    \label{fig:RCM_figure}
\end{figure}

\textbf{Binary Block Masking with Reverse Cuthill-McKee (RCM)} In highly sparse scenarios, processing entire blocks due to a single non-zero entry can lead to inefficiencies, as every block may be required, but only filled with a few non-sparse entries. To mitigate this, we can make use of the Reverse Cuthill-McKee (RCM) \cite{Cuthill1969RCM} algorithm as a preprocessing step. RCM is a well-established graph reordering technique that minimizes the bandwidth of sparse matrices by treating them as connected graphs, effectively increasing the matrix's \textit{"fill-in"}.

While standard Binary Block Masking provides a performance improvement, RCM significantly amplifies this benefit in cases of extreme sparsity. By reorganizing the matrix structure, RCM reduces the number of blocks that must be processed. Figure \ref{fig:RCM_Mask} illustrates a case where RCM reduces the number of blocks by 50\%. Figure \ref{fig:RCM_perf} shows results on a synthetic mask where RCM preprocessing reduces the number of blocks by 90\%, resulting in a significant performance gain. However, it is important to note that the extent of the performance gain is dependent on both the sparsity and the structure of the mask.

Although we focus on the RCM algorithm in this work, other reordering techniques that increase matrix \textit{"fill-in"} could also yield similar improvements in performance when working with sparse matrices. This approach is especially advantageous when the sparsity pattern of the attention mask is either known in advance or can be efficiently computed.

\section{Experiments}
\label{sec:Experiments}
In the following section, we evaluate the performance of our "Binary Block Masking" algorithm using attention masks from three real-world use cases: (1) tree attention masks for Speculative Decoding in MEDUSA\cite{cai2024medusa}, (2) masks from packed finetuning\cite{raffel2020packing} of an LLM on the ALPACA dataset\cite{alpaca}, and (3) fixed masks from the LongFormer paper\cite{Beltagy2020Longformer}.

\subsection{Experimental Setup}
\textbf{Evaluation Metrics} All experiments report total runtime (forward and backward pass) averaged over 100 runs, with a batch size of 4 and 32 attention heads. The \texttt{BLOCKSIZE} for Flash Attention\cite{NEURIPS2022_FlashAttn1, dao2023flashattention2}is fixed to $(128,32)$, which is fine-tuned to achieve optimal performance on our hardware configuration.

\textbf{Baselines} We compare against two baselines: the Triton implementation\cite{tillet2019triton} of Flash Attention\cite{NEURIPS2022_FlashAttn1, dao2023flashattention2} and a naive masking algorithm that applies masks block by block, as previously described. Although Flash Attention is mask-unaware (and hence does not actually produce a correct result), it serves as a useful baseline to display performance gains with our method. We use Flash Attention's Triton implementation (instead of the CUDA implementation \cite{cuda}) for fairer comparisons, as Triton introduces its own overhead.

\textbf{Hardware and Precision} Experiments were conducted on an RTX 3060 (6GB memory) using \texttt{bfloat16} precision for query, key and value vectors. For a more detailed analysis on different hardware, please consult Appendix \ref{sec:HW_perfs}.

\subsection{Preprocessing} We implement a Triton kernel to preprocess the attention mask into the binary block mask format. The preprocessing runtime is comparable to the forward pass of a single attention head. However, this overhead is minimal as the Flash Attention runtime is dominated by the backward pass. Additionally, preprocessing is done once and shared across multiple heads and transformer layers. The following table provides runtime comparisons for various mask dimensions.
\begin{table}[H]
\centering
\caption{Binary Block Preprocessing Time vs Total Runtimes of Flash Attention Algorithm For Varying Head and Batch Sizes}
\small
\begin{tabular}{c c c c c}
\toprule
\textbf{Mask Size} & \textbf{Prepro. Time (ms)} & \textbf{Forward Pass (ms)} & \textbf{Total Runtime (ms)} & \textbf{Total Runtime (ms)} \\
                   &                                 &  \textbf{(B=1, H=1)}    & \textbf{(B=1, H=1)}         & \textbf{(B=4, H=32)}        \\ \midrule
4096   & 0.176                           & 0.306                       & 1.348                       & 96.076                      \\ 
8192   & 0.687                           & 0.886                       & 3.986                       & 377.625                     \\ 
16384 & 2.598                           & 2.935                       & 13.239                      & 1524.763                    \\ 
\bottomrule
\end{tabular}
\end{table}
\normalsize

\subsection{Evaluation on MEDUSA Tree Masks}
\begin{figure}
    \centering
    \begin{subfigure}[c]{0.21\textwidth}
        \centering
        \includegraphics[width=\textwidth]{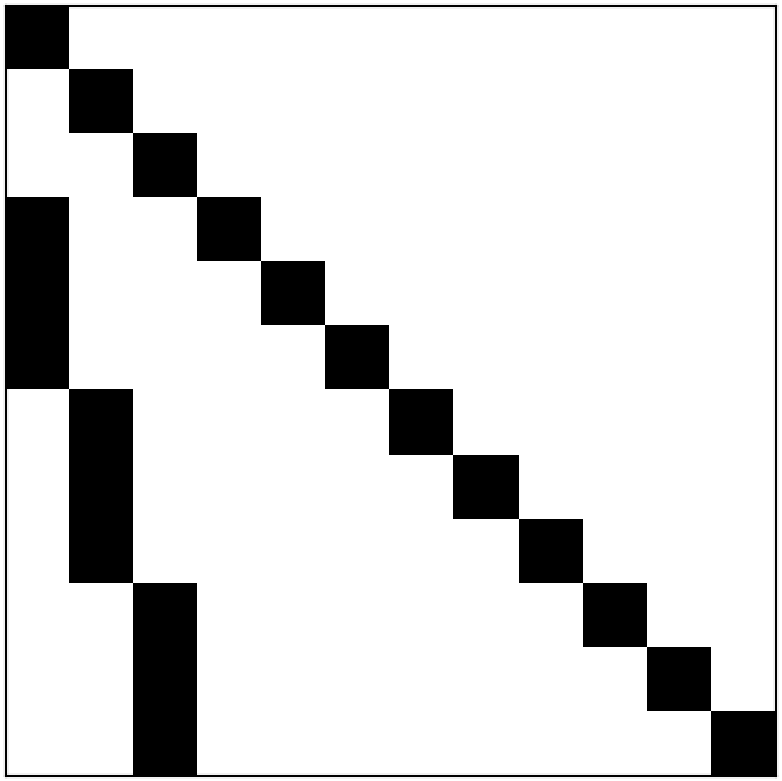}
        \caption{Tree Attention Mask($K$=3, $s_k$=3)}
        \label{fig:MEDUSA_MASKS}
    \end{subfigure}
    \begin{subfigure}[c]{0.78\textwidth}
        \centering
        \includegraphics[height = 5cm, width=0.9\textwidth]{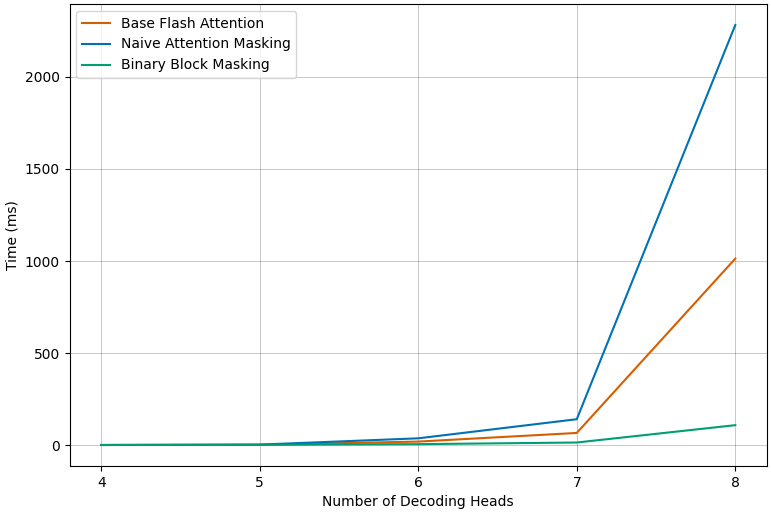}
        \caption{Performance with varying Heads and fixed Candidates($s_k$=3)}
        \label{fig:MEDUSA_HEADS}
        \includegraphics[height = 5cm, width=0.9\textwidth]{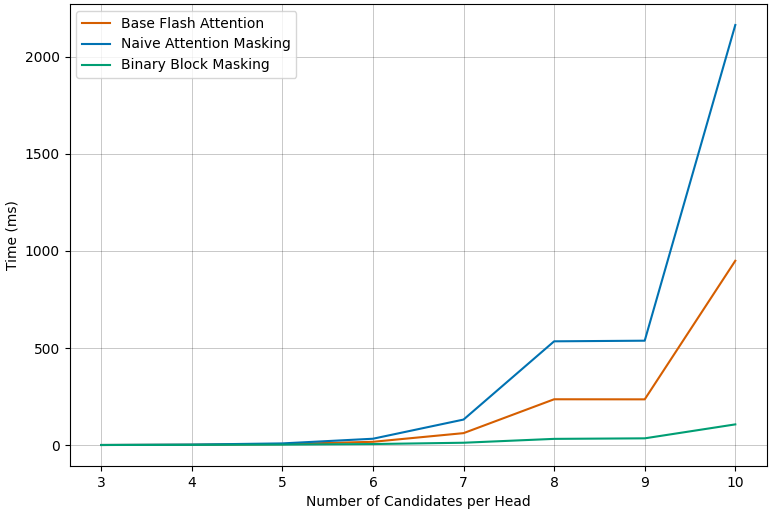}
        \caption{Performance with varying Candidates and fixed Heads($K$=4)}
        \label{fig:MEDUSA_CANDIDATES}
    \end{subfigure}
    \caption{Mask Visualization and Performance Comparison for MEDUSA tree mask}
    \label{fig:MEDUSA_COMBINED}
\end{figure}
Figure \ref{fig:MEDUSA_COMBINED} shows the results for tree masks derived from the MEDUSA architecture\cite{cai2024medusa}, a speculative decoding approach that speeds up inference by using multiple heads to predict future tokens in parallel. MEDUSA generates several candidate tokens per head, verified through a tree-based attention mechanism. We evaluate performance gains from using these tree attention masks.

The size of a MEDUSA tree mask is given by $\sum_{k=1}^{K} \prod_{i=1}^{k} s_i$, where $K$ is the number of decoding heads and $s_k$ is the number of candidates per head. Our current configuration uses 4 heads with 4 candidates each, resulting in a 340x340 attention matrix. At this scale, Flash Attention shows no significant speedup, so we use the native PyTorch implementation as baseline.

Our experiments demonstrate scalability with minimal computational overhead, allowing more decoding heads and candidates per head. Figures \ref{fig:MEDUSA_HEADS}, and \ref{fig:MEDUSA_CANDIDATES} show two scenarios: one with fixed decoding heads($K = 4$) and varying number of candidates($s_k$), and another with fixed number of candidates per-head($s_k$=3) and varying decoding heads($K$). In both, our method outperforms base Flash Attention, showing promise for applications requiring longer sequences and greater variety.

\subsection{Evaluation on packed ALPACA finetuning}
\begin{figure}
    \centering
    \begin{subfigure}[c]{0.21\textwidth}
        \centering
        \includegraphics[width=\textwidth]{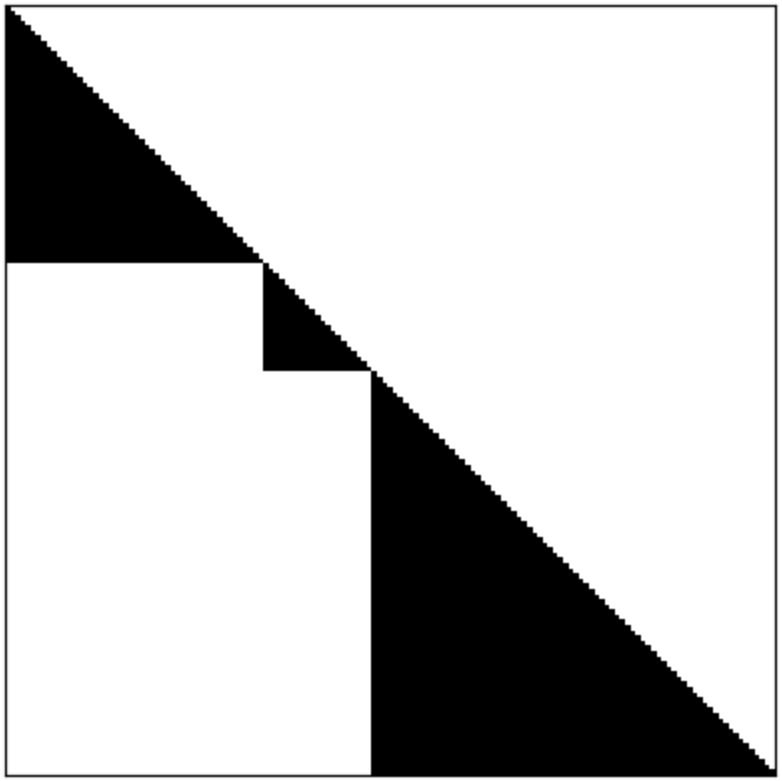}
        \caption{Packed Sequential Mask}
        \label{fig:SEQ_MASK}
    \end{subfigure}
    \begin{subfigure}[c]{0.75\textwidth}
        \centering
        \includegraphics[height = 5cm, width=0.9\textwidth]{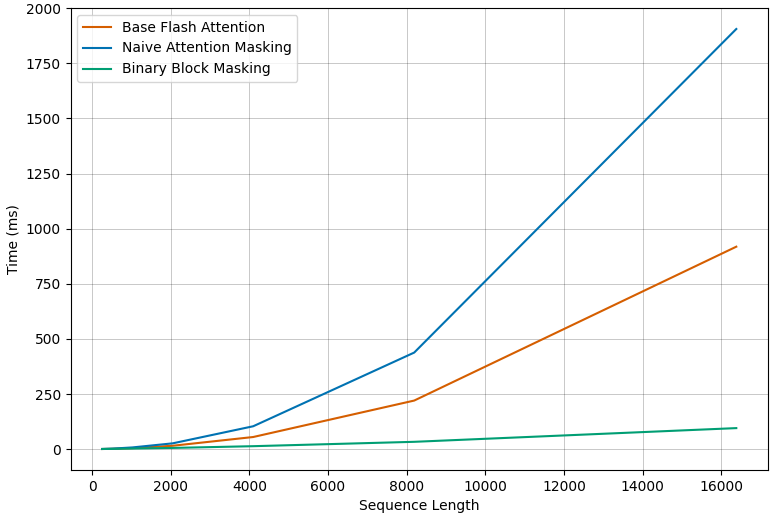}
        \caption{Performance comparison on Sequential Mask}
        \label{fig:SEQ_Perf}
    \end{subfigure}
    \hfill
    \begin{subfigure}[c]{0.21\textwidth}
        \centering
        \includegraphics[width=\textwidth]{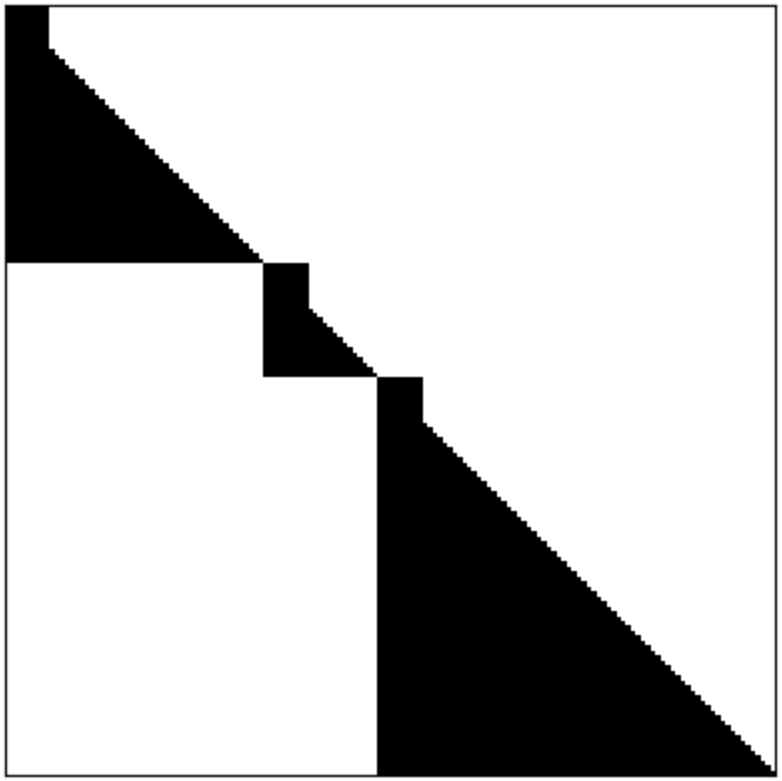}
        \caption{Packed Input-Bidirectional Mask}
        \label{fig:INPUT_BI_MASK}
    \end{subfigure}
    \begin{subfigure}[c]{0.75\textwidth}
        \centering
        \includegraphics[height = 5cm, width=0.9\textwidth]{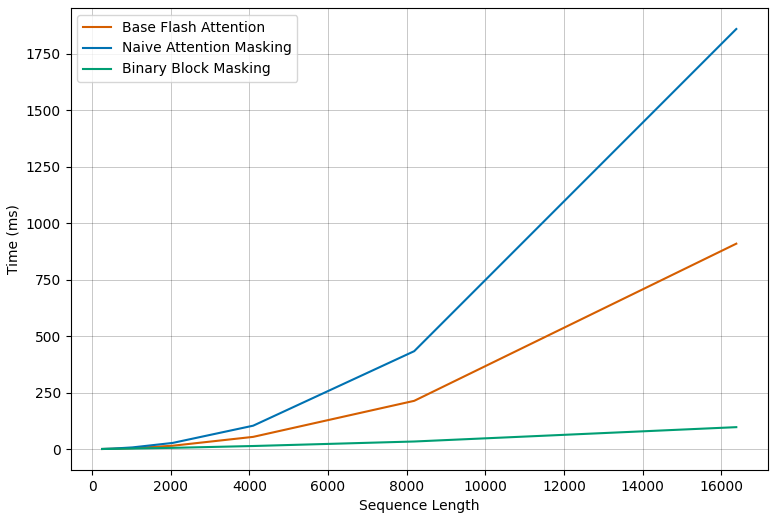}
        \caption{Performance comparison on Input-Bidirectional Mask}
        \label{fig:INPUT_BI_Perf}
    \end{subfigure}
    \caption{Mask Visualization and Performance Comparison for packed ALPACA dataset}
    \label{fig: ALPACA_PERF}
\end{figure}
ALPACA\cite{alpaca} is an instruction fine-tuning dataset comprising input, instruction, and output components. We combine the input and instruction into a single input and pack multiple sequences into longer ones, using masks to prevent cross-sequence attention. Figure \ref{fig: ALPACA_PERF} demonstrates our algorithm's performance on two ALPACA packing approaches: (1) \textbf{Sequential Masks}: Input and output are concatenated, with tokens attending only to previous tokens. And (2) \textbf{Input-Bidirectional Mask}: Input tokens attend to each other, while output tokens attend to input and preceding output tokens. Both masks consist of contiguous non-zero blocks, enabling the use of our \texttt{Dense BinBlkMsk} algorithm variant.

Packing improves upon padding, where shorter sequences in a batch are extended with a special token to match the longest sequence's length. While padding wastes GPU resources, it's more compatible with Flash Attention. Our experiments show that our algorithms efficiently combine the benefits of Flash Attention with packing, resulting in approximately a 9x performance increase in both cases.

\subsection{Evaluation on LongFormer Attention Masks}
\begin{figure}
    \centering
    \begin{subfigure}[c]{0.21\textwidth}
        \centering
        \includegraphics[width=\textwidth]{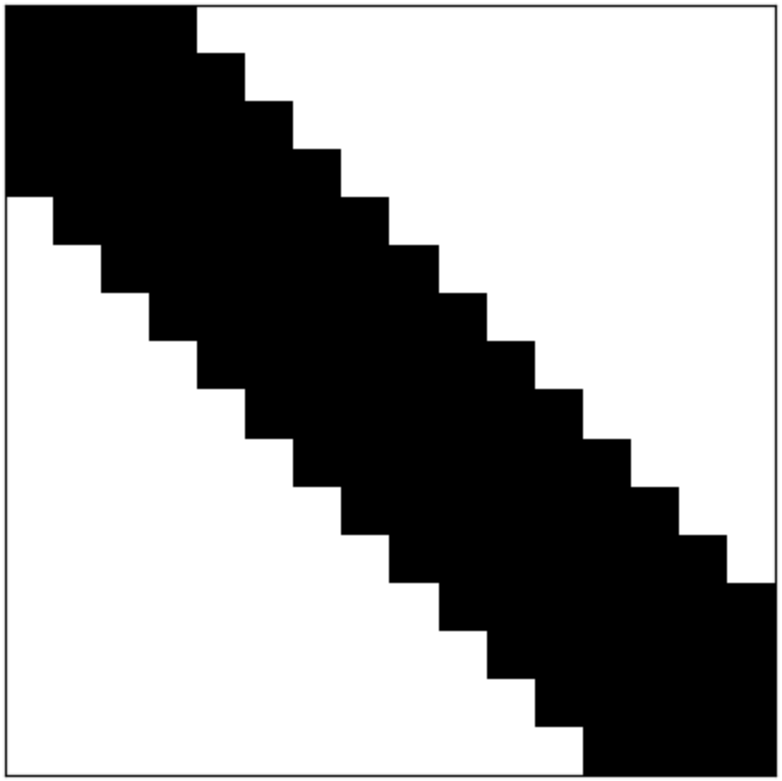}
        \caption{Longformer - Windowed Mask}
    \end{subfigure}
    \begin{subfigure}[c]{0.75\textwidth}
        \centering
        \includegraphics[height = 4cm, width=0.9\textwidth]{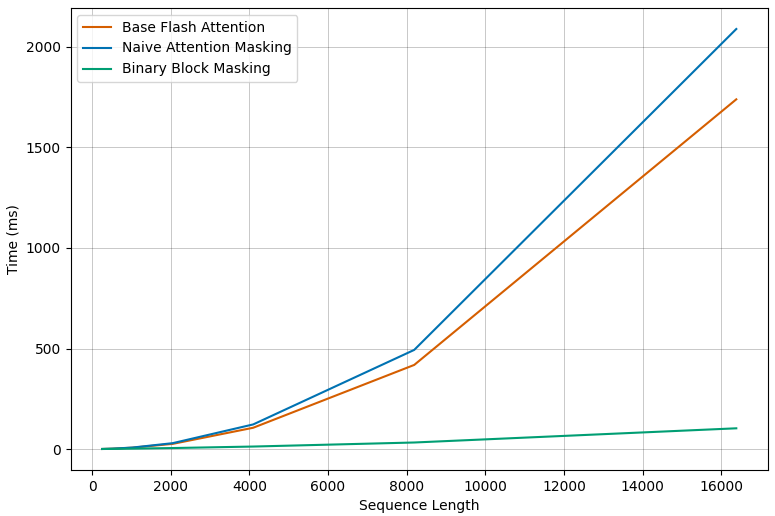}
        \caption{Performance comparison on Windowed Mask}
    \end{subfigure}
    \hfill
    \begin{subfigure}[c]{0.21\textwidth}
        \centering
        \includegraphics[width=\textwidth]{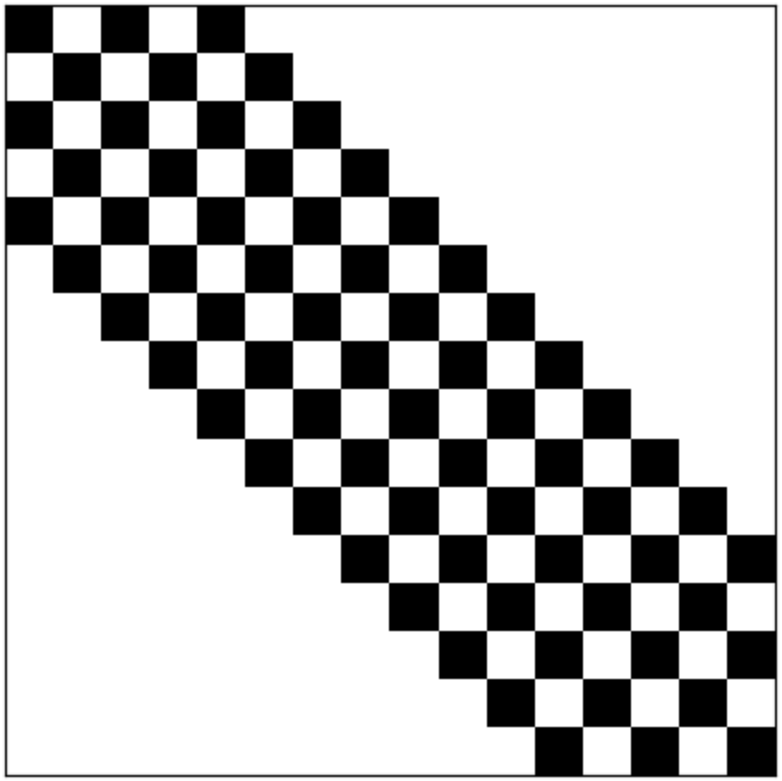}
        \caption{Longformer - Dilated Mask}
    \end{subfigure}
    \begin{subfigure}[c]{0.75\textwidth}
        \centering
        \includegraphics[height = 4cm, width=0.9\textwidth]{Images/LongFormer/Dilated_Perf.png}
        \caption{Performance comparison on Dilated Mask}
    \end{subfigure}
    \hfill
    \begin{subfigure}[c]{0.21\textwidth}
        \centering
        \includegraphics[width=\textwidth]{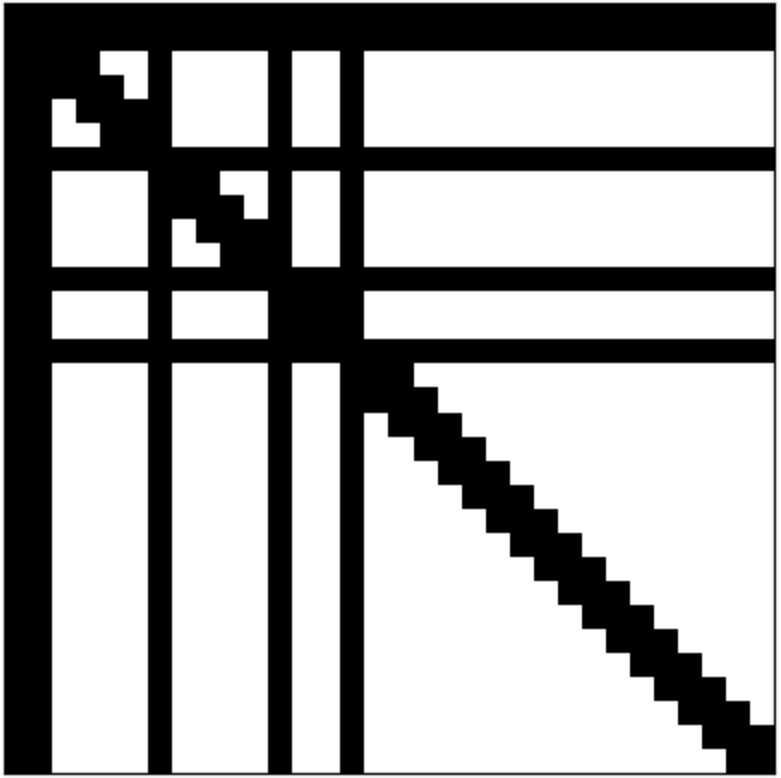}
        \caption{Longformer - Global Mask}
    \end{subfigure}
    \begin{subfigure}[c]{0.75\textwidth}
        \centering
        \includegraphics[height = 4cm, width=0.9\textwidth]{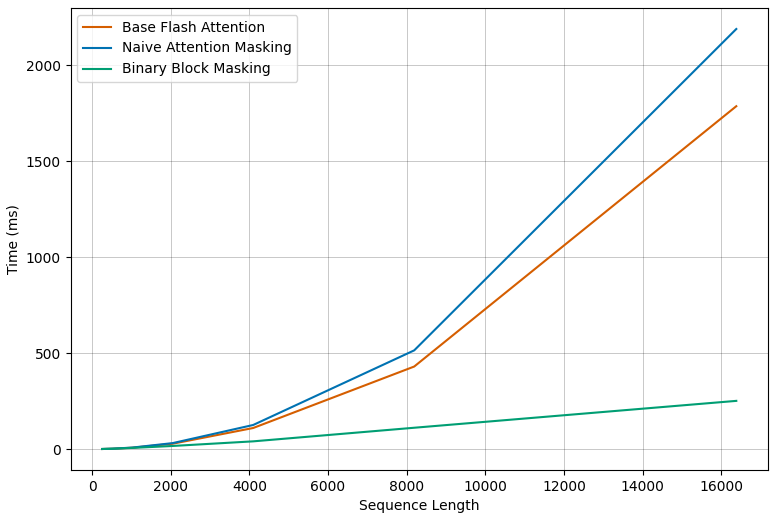}
        \caption{Performance comparison on Global Mask}
    \end{subfigure}
    \caption{Mask Visualization and Performance Comparison for Longformer}
    \label{fig:LONGFORMER_PERF}
\end{figure}
We evaluate our algorithm across three LongFormer\cite{Beltagy2020Longformer} masking patterns: Windowed, Dilated, and Global. For the Windowed mask, we use the \texttt{Dense BinBlkMsk}, while the base algorithm is used for the other two masks. The performance results, along with the corresponding masks, are presented in Figure \ref{fig:LONGFORMER_PERF}. 

Our findings demonstrate that our method consistently outperforms the baseline Flash Attention algorithm across all three masks. Moreover, these results highlight that the performance improvements gained through sparsity are complementary to those achieved via FlashAttention. By leveraging our mask-aware algorithm, we can construct more efficient architectures capable of processing longer sequences with faster processing.

\section{Discussion, Limitations, and Future Work}
\label{sec:Limitations_and_Future}
Our experimental results demonstrate that the proposed algorithm effectively leverages the mask structure to achieve notable speedups, though the extent of these improvements is highly dependent on the mask's fill pattern. Specifically, as the mask becomes increasingly filled, the processing time grows correspondingly. However, even in extreme cases—such as a fully dense mask (composed entirely of ones) or a causal mask—our algorithm maintains performance that is competitive with Flash Attention\cite{NEURIPS2022_FlashAttn1, dao2023flashattention2}. This is primarily due to the optimization introduced by the \texttt{Dense BinBlkMsk} representation, which ensures that the only additional overhead compared to the base implementation is the reading of binary values for the Binary Block Mask (Appendix \ref{sec:Ones_and_Causal}).

One scenario where our method may exhibit slower performance than Flash Attention arises when the mask is nearly full but consists of non-contiguous non-zero blocks. In such cases, Flash Attention may outperform our algorithm in raw speed. However, it is important to note that Flash Attention is mask-agnostic, meaning the output it generates does not inherently respect the mask constraints. To apply masking with Flash Attention, a post-processing step is required, which can negate its initial speed advantage and make the overall process slower.

For future work, a key direction would be migrating our custom Triton kernels\cite{tillet2019triton} to CUDA\cite{cuda} and integrating them with the state-of-the-art Flash Attention framework to further enhance performance. Additionally, it is crucial to evaluate the algorithm on real-world, end-to-end tasks rather than focusing solely on isolated mask cases. Another avenue is exploring alternatives to Reverse Cuthill-McKee\cite{Cuthill1969RCM} for increasing matrix fill-in, enabling us to extend the approach to more general sparse matrices, including asymmetric matrices or matrices without underlying connected graph structures.

\section{Conclusion}
\label{sec:Conclusion}
In this paper, we present our implementation of \textbf{Binary Block Masking} that optimizes Flash Attention for partially filled attention masks. By preprocessing the mask blocks in parallel, our method identifies non-zero blocks and restricts attention computation to these blocks, enhancing efficiency. This approach is both general-purpose and easy to integrate, making it broadly applicable. Our experiments demonstrate substantial performance improvements over base Flash Attention and highlight the versatility of our method across diverse applications. We will release our Triton kernels and mask-processing code to encourage further research in this area.

\begin{ack}
    We appreciate Madhavi Sen for her help with formatting of figures and tables, Anna Manasyan for proof reading, and Dr. Philipp Hennig for inspiring this collaboration. Dr. Jonas Geiping acknowledges the support of the Hector Foundation II, the Max Planck Society, and the T\"ubingen AI Center in T\"ubingen, Germany. Lastly, Agniv Sharma's work is supported by his hopes and dreams.
\end{ack}
\bibliographystyle{plainnat}
\bibliography{references}

\newpage
\appendix

\section*{\centering APPENDIX}
\vspace{10pt}
\section{Preprocessing BinBlkMat: Method Description and Analysis}
\label{sec:Preprocessing_BinBlkMat}
We explored several different kernel implementations to compute \texttt{BinBlkMat}. In the following descriptions, \texttt{N} refers to the sequence length or the dimension of the original mask, \texttt{BLOCKSIZE\_I} is the block size along the rows, and \texttt{BLOCKSIZE\_J} is the block size along the columns. The output \texttt{BinBlkMat} has a shape of $N//\texttt{BLOCKSIZE\_I} \times N//\texttt{BLOCKSIZE\_J}$. The approaches we tested are as follows:
\begin{enumerate}

\item \textbf{Convolutional approach}: We used \texttt{torch.nn.functional.conv2d} with a stride of \texttt{BLOCKSIZE}.

\item \textbf{Direct kernel grid}: We launch a grid of $\texttt{N} \times \texttt{N}$ kernels, where each kernel \texttt{K\textsubscript{ij}} checks whether the corresponding entry \texttt{(i,j)} in the original mask was 0 or 1. If the value was 1, it assigned 1 to the index \texttt{(i//BLOCKSIZE\_I, j//BLOCKSIZE\_J)} in \texttt{BinBlkMat}.

\item \textbf{Block-based summation kernel}: We launched a grid of size $\texttt{N//BLOCKSIZE\_I} \times \texttt{N//BLOCKSIZE\_J} $, where each kernel operated over a block of size $\texttt{BLOCKSIZE\_I} \times \texttt{BLOCKSIZE\_J}$. The kernel computed the sum of the block elements, assigning a 1 to the corresponding block in \texttt{BinBlkMat} if the sum was greater than zero.

\item \textbf{In-built Torch operations}: Similar to the third approach, but using native PyTorch\cite{pytorch} operations for block summation and assignment.

\end{enumerate}
Among these methods, the third approach provided the best performance. Moreover, the sum output from this approach is also used to calculate \texttt{total\_ones} and \texttt{offset} for Dense Binary Block Masking cases, enabling the efficient computation of two outputs with minimal overhead. The runtime for the third approach is reported in Table \ref{table:Prepro_BinBlk}

\begin{table}[H]
  \caption{Preprocessing runtime for BinBlkMat}
  \label{table:Prepro_BinBlk}
  \centering
  \begin{tabular}{ccc}
    \toprule
    \textbf{Mask Size} & \textbf{BinBlkMat (ms)} & \textbf{BinBlkMat with total\_ones and offset (ms)} \\
    \midrule
    256 & 0.06 & 0.12 \\
    512 & 0.06 & 0.12 \\
    1024 & 0.06 & 0.12 \\
    2048 & 0.08 & 0.14 \\
    4096 & 0.19 & 0.26 \\
    8192 & 0.67 & 0.73 \\
    16384 & 2.92 & 2.96 \\
    \bottomrule
\end{tabular}
\end{table}

\section{Performance Comparison On Different GPUs}
\label{sec:HW_perfs}
In the main paper, we use an RTX 3060 GPU with 6GB of VRAM for benchmark evaluations. In this section, we extend the analysis to include runtimes on four additional GPUs: A100, H100, T4, and RTX 6000 Ada Generation. We begin by presenting the runtimes of our preprocessing kernels, followed by the performance results on the Input-Bidirectional ALPACA mask\cite{alpaca}.

\subsection{Preprocessing Runtime Comparisons}

Tables \ref{table:Prepro_A100},\ref{table:Prepro-H100},\ref{table:Prepro-T4},\ref{table:Prepro-Ada6000} give the performance comparison for A100, H100, T4, and RTX 6000 Ada Generation respectively. Apart from T4, the preprocessing times decrease even further for more powerful GPUs.

\begin{table}
  \caption{Preprocessing runtime for BinBlkMat: RTX 3060 vs. A100}
  \label{table:Prepro_A100}
  \centering
  \begin{tabular}{lcccc}
    \toprule
    \textbf{Seq. Length} & \multicolumn{2}{c}{\textbf{RTX 3060}} & \multicolumn{2}{c}{\textbf{A100}} \\
    \cmidrule(lr){2-3} \cmidrule(lr){4-5}
                    & \textbf{BinBlkMsk}  & \textbf{BinBlkMsk with}   & \textbf{BinBlkMsk}  & \textbf{BinBlkMsk with} \\
                    &&\textbf{total\_ones and offset}&&\textbf{total\_ones and offset}\\
    \midrule
    256             & 0.06  & 0.12  & 0.07  & 0.16  \\
    512             & 0.06  & 0.12  & 0.07  & 0.16  \\
    1024            & 0.06  & 0.12  & 0.07  & 0.16  \\
    2048            & 0.08  & 0.15  & 0.07  & 0.16  \\
    4096            & 0.19  & 0.25  & 0.10  & 0.19  \\
    8192            & 0.67  & 0.73  & 0.20  & 0.29  \\
    16384           & 2.92  & 2.96  & 0.64  & 0.73  \\
    \bottomrule
  \end{tabular}
\end{table}

\begin{table}
  \caption{Preprocessing runtime for BinBlkMat: RTX 3060 vs. H100}
  \label{table:Prepro-H100}
  \centering
  \begin{tabular}{lcccc}
    \toprule
    \textbf{Seq. Length} & \multicolumn{2}{c}{\textbf{RTX 3060}} & \multicolumn{2}{c}{\textbf{H100}} \\
    \cmidrule(lr){2-3} \cmidrule(lr){4-5}
                    & \textbf{BinBlkMsk}  & \textbf{BinBlkMsk with}   & \textbf{BinBlkMsk}  & \textbf{BinBlkMsk with} \\
                    &&\textbf{total\_ones and offset}&&\textbf{total\_ones and offset}\\
    \midrule
    256             & 0.06  & 0.12  & 0.05  & 0.11  \\
    512             & 0.06  & 0.12  & 0.05  & 0.11  \\
    1024            & 0.06  & 0.12  & 0.05  & 0.11  \\
    2048            & 0.08  & 0.15  & 0.05  & 0.11  \\
    4096            & 0.19  & 0.25  & 0.07  & 0.13  \\
    8192            & 0.67  & 0.73  & 0.15  & 0.21  \\
    16384           & 2.92  & 2.96  & 0.47  & 0.53  \\
    \bottomrule
  \end{tabular}
\end{table}

\begin{table}
  \caption{Preprocessing runtime for BinBlkMat: RTX 3060 vs. T4}
  \label{table:Prepro-T4}
  \centering
  \begin{tabular}{lcccc}
    \toprule
    \textbf{Seq. Length} & \multicolumn{2}{c}{\textbf{RTX 3060}} & \multicolumn{2}{c}{\textbf{T4}} \\
    \cmidrule(lr){2-3} \cmidrule(lr){4-5}
                    & \textbf{BinBlkMsk}  & \textbf{BinBlkMsk with}   & \textbf{BinBlkMsk}  & \textbf{BinBlkMsk with} \\
                    &&\textbf{total\_ones and offset}&&\textbf{total\_ones and offset}\\
    \midrule
    256             & 0.06  & 0.12  & 0.11  & 0.21  \\
    512             & 0.06  & 0.12  & 0.11  & 0.24  \\
    1024            & 0.06  & 0.12  & 0.11  & 0.23  \\
    2048            & 0.08  & 0.15  & 0.15  & 0.27  \\
    4096            & 0.19  & 0.25  & 0.30  & 0.41  \\
    8192            & 0.67  & 0.73  & 1.05  & 1.17  \\
    16384           & 2.92  & 2.96  & 4.66  & 4.80  \\
    \bottomrule
  \end{tabular}
\end{table}

\begin{table}
  \caption{Preprocessing runtime for BinBlkMat: RTX 3060 vs. RTX 6000 Ada}
  \label{table:Prepro-Ada6000}
  \centering
  \begin{tabular}{lcccc}
    \toprule
   \textbf{Seq. Length} & \multicolumn{2}{c}{\textbf{RTX 3060}} & \multicolumn{2}{c}{\textbf{RTX 6000 Ada}} \\
    \cmidrule(lr){2-3} \cmidrule(lr){4-5}
                    & \textbf{BinBlkMsk}  & \textbf{BinBlkMsk with}   & \textbf{BinBlkMsk}  & \textbf{BinBlkMsk with} \\
                    &&\textbf{total\_ones and offset}&&\textbf{total\_ones and offset}\\
    \midrule
    256             & 0.06  & 0.12  & 0.06  & 0.13  \\
    512             & 0.06  & 0.12  & 0.06  & 0.13  \\
    1024            & 0.06  & 0.12  & 0.06  & 0.13  \\
    2048            & 0.08  & 0.15  & 0.06  & 0.14  \\
    4096            & 0.19  & 0.25  & 0.12  & 0.19  \\
    8192            & 0.67  & 0.73  & 0.41  & 0.49  \\
    16384           & 2.92  & 2.96  & 1.48  & 1.54  \\
    \bottomrule
  \end{tabular}
\end{table}
\newpage
\subsection{ALPACA Input-Bidirectional Mask Runtime Comparison}
The tables \ref{table:ALPACA_A100},\ref{table:ALPACA_H100},\ref{table:ALPACA_T4},\ref{table:ALPACA_ADA6000} present the runtime comparisons for the A100, H100, T4, and RTX 6000 Ada Generation GPUs, respectively. We employed a \texttt{BLOCKSIZE} of $(128,32)$, originally fine-tuned for the RTX 3060. While these performance metrics are not optimized for the specific hardware configurations tested, the results still clearly demonstrate the advantages of using the Binary Block Masking algorithm. Further optimization could potentially yield even better runtimes across different hardware.
\begin{table}[H]
  \caption{Performance comparison on ALPACA Input-Bidirectional mask: RTX 3060 vs. A100}
  \label{table:ALPACA_A100}
  \centering
  \begin{tabular}{lcccc}
    \toprule
    \textbf{Seq. Length} & \multicolumn{2}{c}{\textbf{RTX 3060}} & \multicolumn{2}{c}{\textbf{A100}} \\
    \cmidrule(lr){2-3} \cmidrule(lr){4-5}
                    & \textbf{Base Flash Attention}  & \textbf{BinBlkMsk}  & \textbf{Base Flash Attention}  & \textbf{BinBlkMsk}  \\
    \midrule
    256             & 0.58  & 0.59  & 0.17  & 0.23  \\
    512             & 1.50  & 1.32  & 0.33  & 0.44  \\
    1024            & 4.45  & 2.64  & 0.86  & 0.80  \\
    2048            & 15.06 & 5.68  & 2.26  & 1.74  \\
    4096            & 54.32 & 13.82 & 7.69  & 4.22  \\
    8192            & 213.92 & 33.80 & 28.38 & 10.01 \\
    16384           & 909.90 & 97.34 & 107.92 & 33.75 \\
    \bottomrule
  \end{tabular}
\end{table}

\begin{table}[H]
  \caption{Performance comparison on ALPACA Input-Bidirectional mask: RTX 3060 vs. H100}
  \label{table:ALPACA_H100}
  \centering
  \begin{tabular}{lcccc}
    \toprule
    \textbf{Seq. Length} & \multicolumn{2}{c}{\textbf{RTX 3060}} & \multicolumn{2}{c}{\textbf{H100}} \\
    \cmidrule(lr){2-3} \cmidrule(lr){4-5}
                    & \textbf{Base Flash Attention}  & \textbf{BinBlkMsk}  & \textbf{Base Flash Attention}  & \textbf{BinBlkMsk}  \\
    \midrule
    256             & 0.58  & 0.59  & 0.19  & 0.16  \\
    512             & 1.50  & 1.32  & 0.18  & 0.24  \\
    1024            & 4.45  & 2.64  & 0.42  & 0.43  \\
    2048            & 15.06 & 5.68  & 1.16  & 0.90  \\
    4096            & 54.32 & 13.82 & 3.82  & 2.16  \\
    8192            & 213.92 & 33.80 & 13.78 & 5.85  \\
    16384           & 909.90 & 97.34 & 50.99 & 18.44 \\
    \bottomrule
  \end{tabular}
\end{table}

\begin{table}[H]
  \caption{Performance comparison on ALPACA Input-Bidirectional mask: RTX 3060 vs. T4}
  \label{table:ALPACA_T4}
  \centering
  \begin{tabular}{lcccc}
    \toprule
    \textbf{Seq. Length} & \multicolumn{2}{c}{\textbf{RTX 3060}} & \multicolumn{2}{c}{\textbf{T4}} \\
    \cmidrule(lr){2-3} \cmidrule(lr){4-5}
                    & \textbf{Base Flash Attention}  & \textbf{BinBlkMsk}  & \textbf{Base Flash Attention}  & \textbf{BinBlkMsk}  \\
    \midrule
    256             & 0.58  & 0.59  & 0.97  & 1.19  \\
    512             & 1.50  & 1.32  & 1.84  & 2.52  \\
    1024            & 4.45  & 2.64  & 5.15  & 3.36  \\
    2048            & 15.06 & 5.68  & 15.95 & 7.57  \\
    4096            & 54.32 & 13.82 & 57.14 & 19.30 \\
    8192            & 213.92 & 33.80 & 215.30 & 49.26 \\
    16384           & 909.90 & 97.34 & 869.90 & 137.02 \\
    \bottomrule
  \end{tabular}
\end{table}

\begin{table}[H]
  \caption{Performance comparison on ALPACA Input-Bidirectional mask: RTX 3060 vs. RTX 6000 Ada}
  \label{table:ALPACA_ADA6000}
  \centering
  \begin{tabular}{lcccc}
    \toprule
   \textbf{Seq. Length} & \multicolumn{2}{c}{\textbf{RTX 3060}} & \multicolumn{2}{c}{\textbf{RTX 6000 Ada}} \\
    \cmidrule(lr){2-3} \cmidrule(lr){4-5}
                    & \textbf{Base Flash Attention}  & \textbf{BinBlkMsk}  & \textbf{Base Flash Attention}  & \textbf{BinBlkMsk}  \\
    \midrule
    256             & 0.58  & 0.59  & 0.14  & 0.26  \\
    512             & 1.50  & 1.32  & 0.28  & 0.29  \\
    1024            & 4.45  & 2.64  & 0.71  & 0.56  \\
    2048            & 15.06 & 5.68  & 2.30  & 1.16  \\
    4096            & 54.32 & 13.82 & 8.03  & 2.61  \\
    8192            & 213.92 & 33.80 & 30.69 & 6.48  \\
    16384           & 909.90 & 97.34 & 113.76 & 18.06 \\
    \bottomrule
  \end{tabular}
\end{table}

\section{Performance Comparison on Causal and All-Ones mask}
\label{sec:Ones_and_Causal}
\begin{figure}[H]
    \centering
    \begin{subfigure}[c]{0.21\textwidth}
        \centering
        \includegraphics[width=\textwidth]{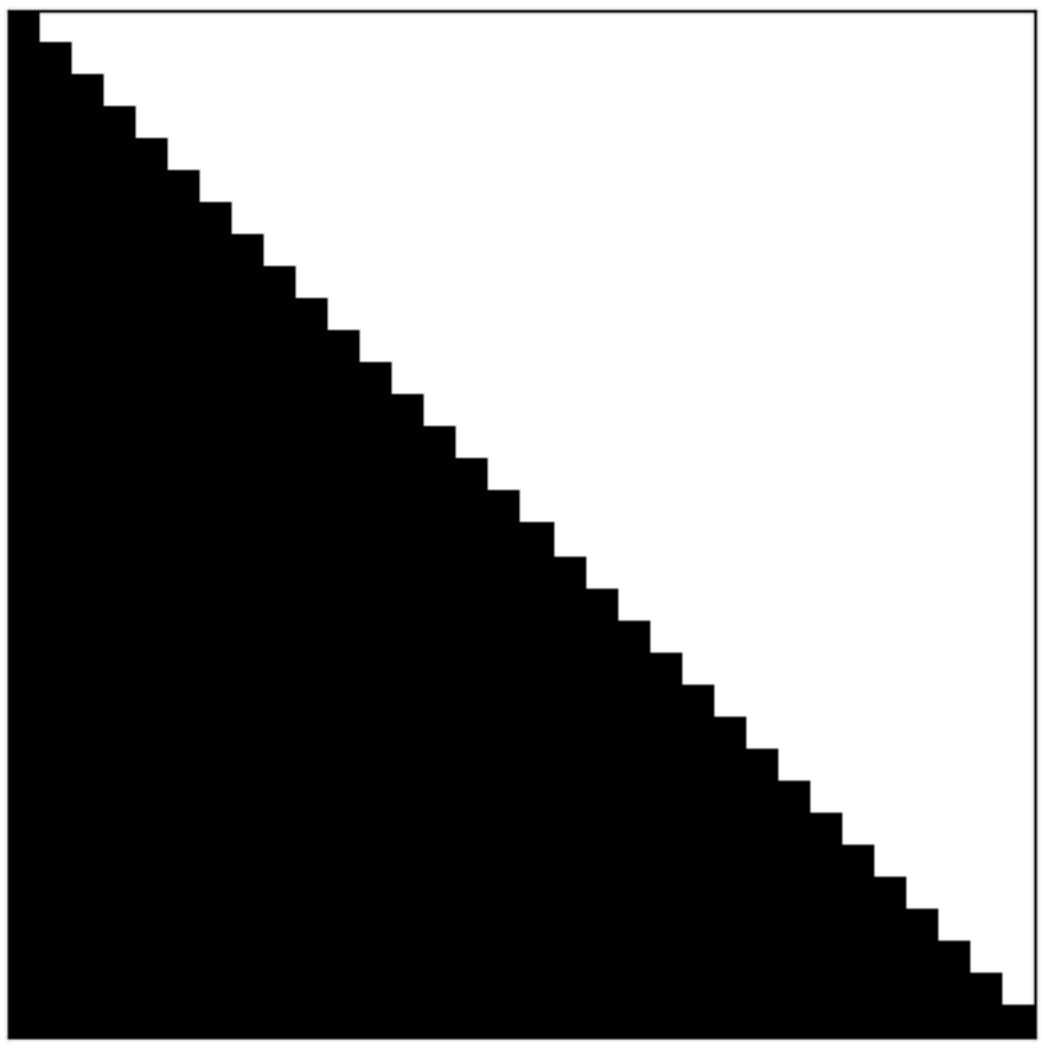}
        \caption{Causal Mask}
    \end{subfigure}
    \begin{subfigure}[c]{0.75\textwidth}
        \centering
        \includegraphics[height = 5cm, width=0.9\textwidth]{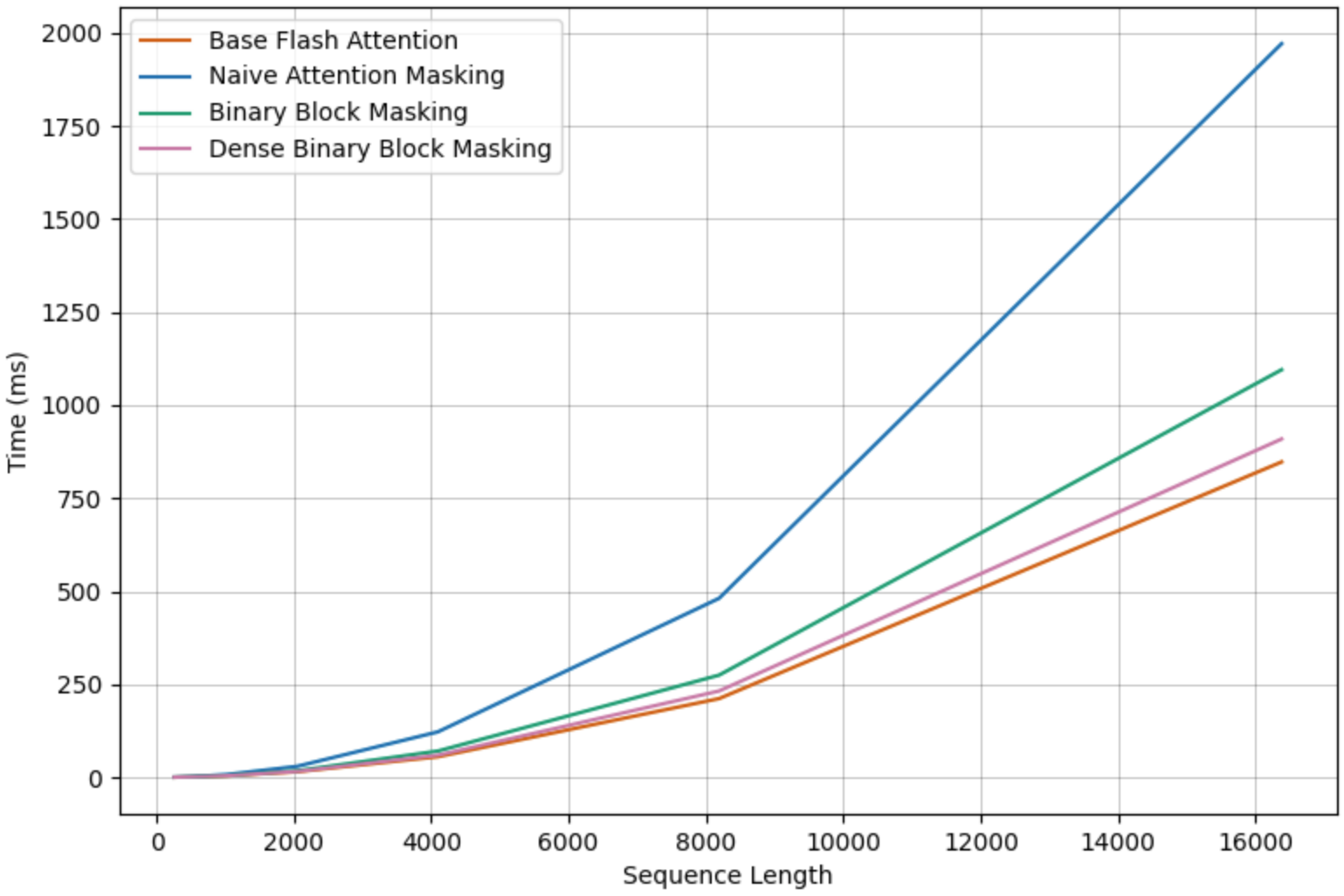}
        \caption{Performance comparison on Causal Mask}
    \end{subfigure}
    \caption{Mask Visualization and Performance Comparison for Causal Masks}
    \label{fig:CAUSAL_PERF}
\end{figure}

\begin{figure}[H]
    \centering
    \begin{subfigure}[c]{0.21\textwidth}
        \centering
        \includegraphics[width=\textwidth]{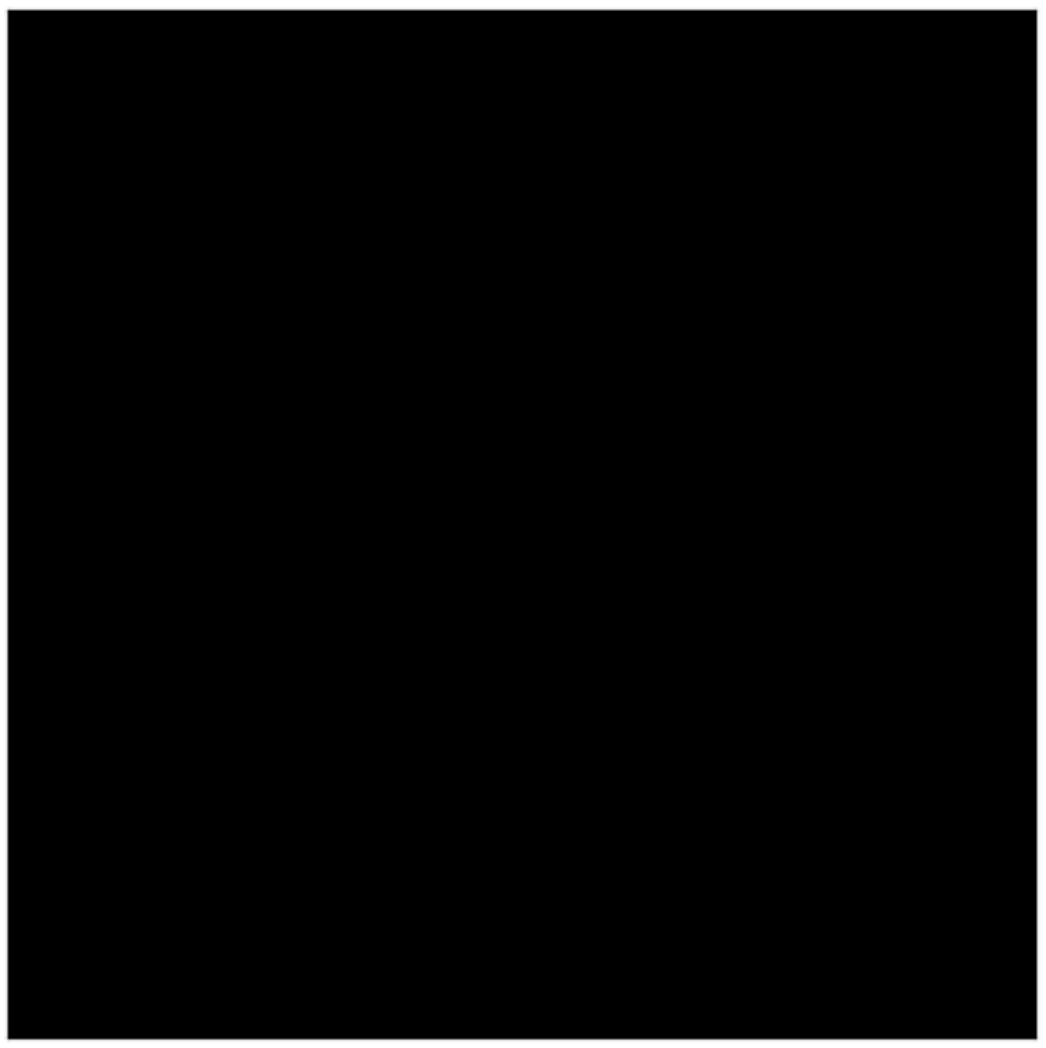}
        \caption{All-Ones Mask}
    \end{subfigure}
    \begin{subfigure}[c]{0.75\textwidth}
        \centering
        \includegraphics[height = 5cm, width=0.9\textwidth]{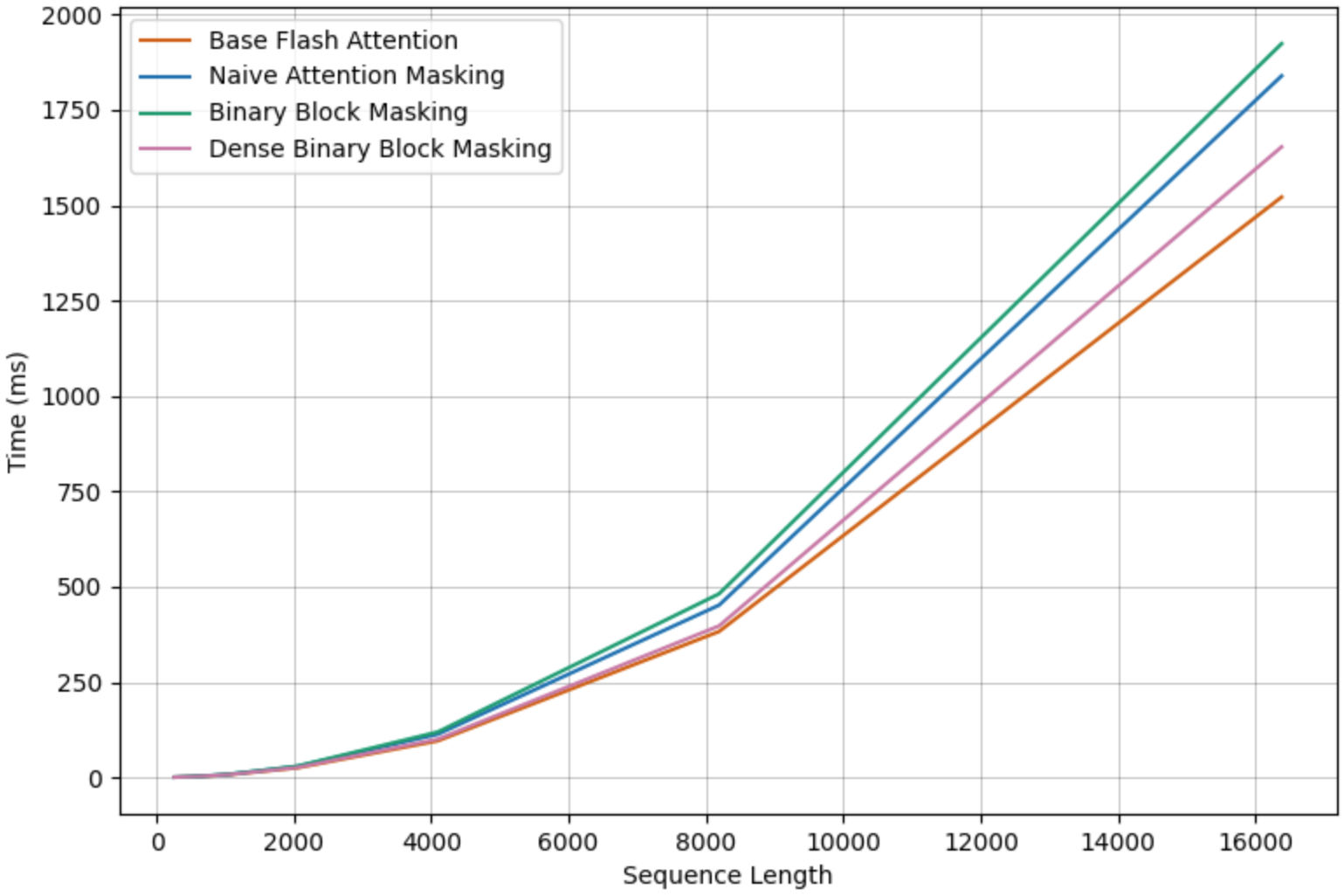}
        \caption{Performance comparison on All-Ones Mask}
    \end{subfigure}
    \caption{Mask Visualization and Performance Comparison for Causal Masks}
    \label{fig:ONES_PERF}
\end{figure}

In this section, we present the performance of our algorithm under two extreme masking conditions: the Causal Mask (Figure \ref{fig:CAUSAL_PERF}) and the All-Ones Mask (Figure \ref{fig:ONES_PERF}). Our approach leverages partial mask fill to optimize computational efficiency. As the mask becomes denser, the performance of the algorithm can degrade. However, even under these extreme cases, our "Dense Binary Block Masking" technique effectively exploits the contiguous structure of the masks, resulting in comparable performance.

\section{Forward and Backward Pass Performance Comparisons}

\begin{figure}[H]
    \centering
    \begin{subfigure}[b]{0.49\textwidth}
        \centering
    \includegraphics[width=\textwidth]{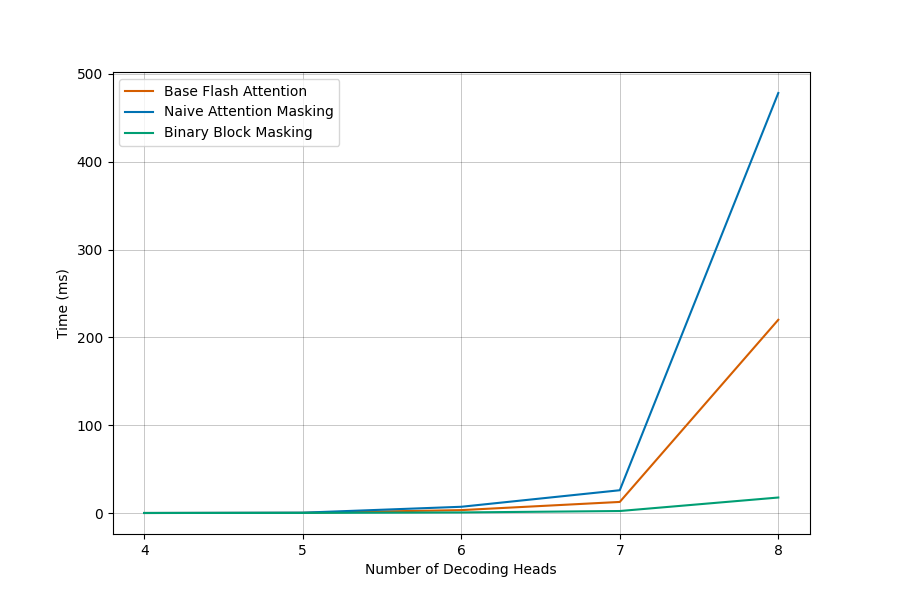}
        \caption{Forward Pass}
    \end{subfigure}
    \hfill
    \begin{subfigure}{0.49\textwidth}
        \centering
    \includegraphics[width=\textwidth]{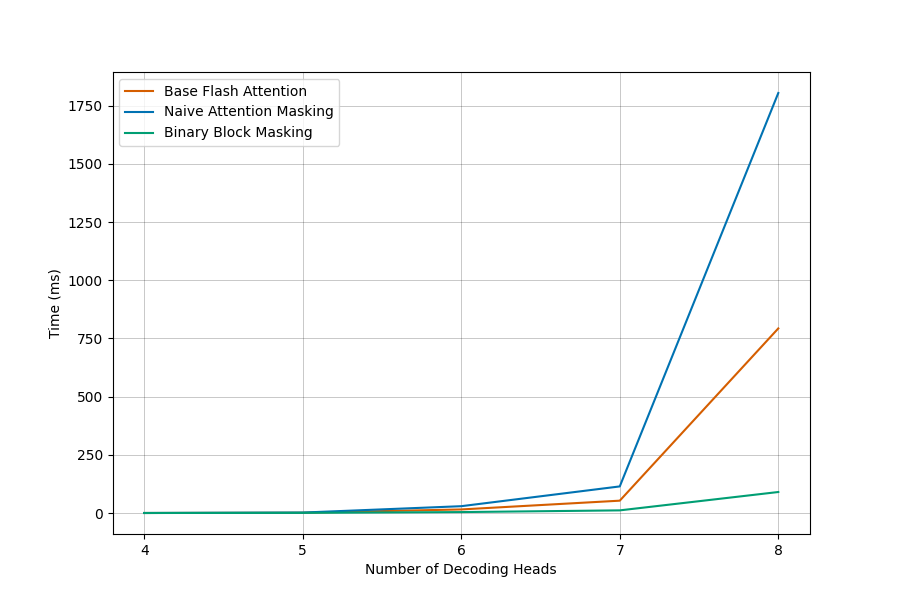}
        \caption{Backward Pass}
    \end{subfigure}
    
    \caption{MEDUSA: Varying Decoding Heads, Number of Candidates Fixed}
    \label{fig:Appendix_Medusa_Heads}
\end{figure}

\begin{figure}[H]
    \centering
    \begin{subfigure}[b]{0.49\textwidth}
        \centering
        \includegraphics[width=\textwidth]{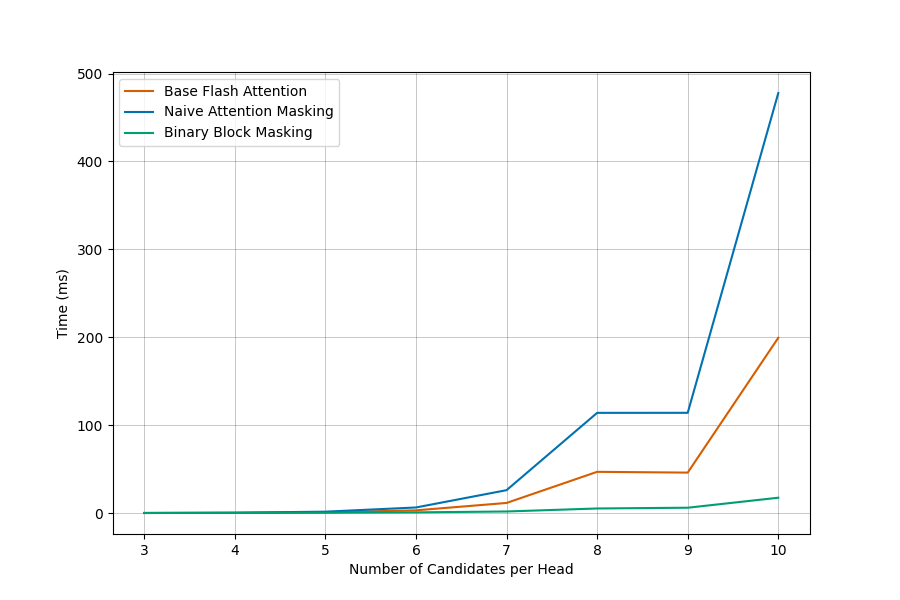}
        \caption{Forward Pass}
    \end{subfigure}
    \hfill
    \begin{subfigure}{0.49\textwidth}
        \centering
        \includegraphics[width=\textwidth]{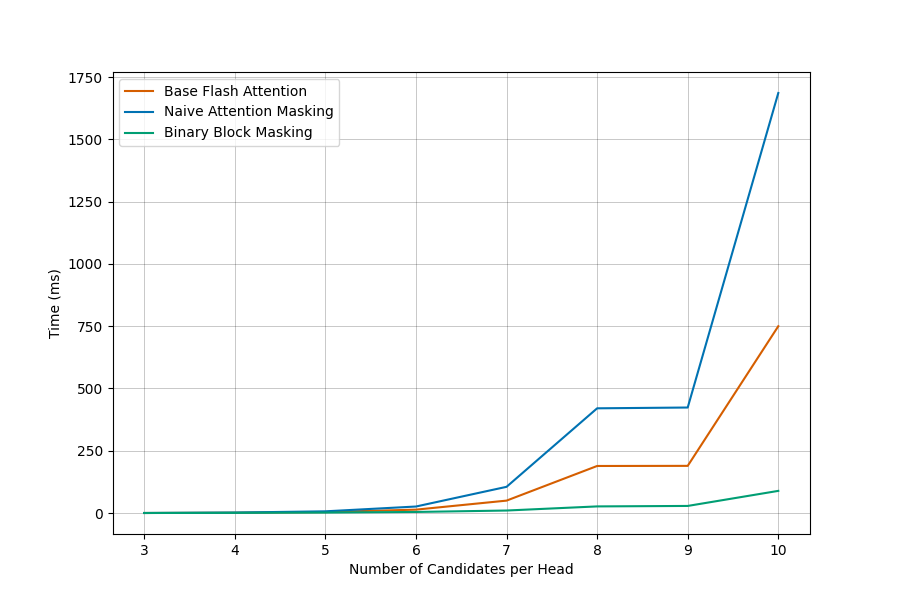}
        \caption{Backward Pass}
    \end{subfigure}
    
    \caption{MEDUSA: Varying Number of Candidates, Fixed Decoding Heads}
    \label{fig:Appendix_Medusa_Predicitions}
\end{figure}

\begin{figure}[H]
    \centering
    \begin{subfigure}[b]{0.49\textwidth}
        \centering
        \includegraphics[width=\textwidth]{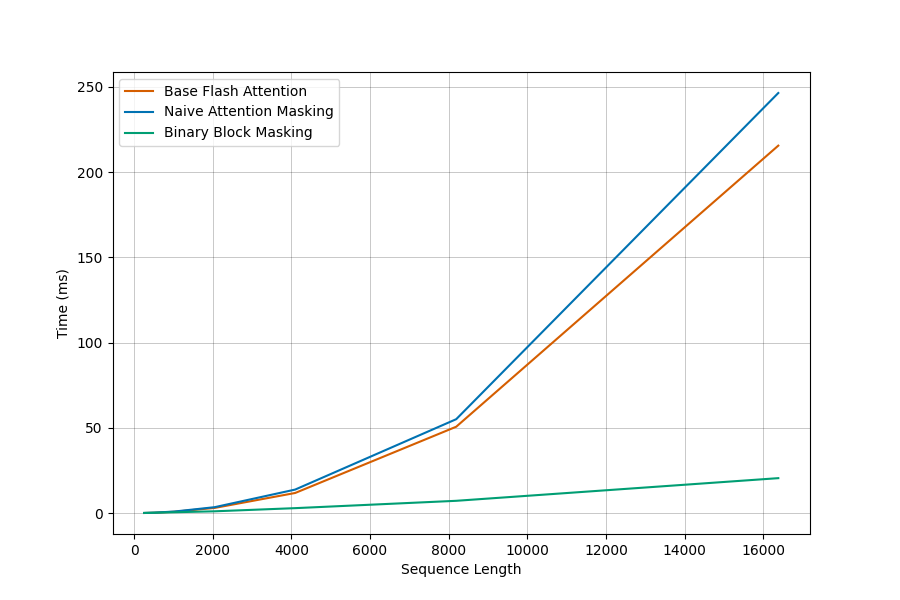}
        \caption{Forward Pass}
    \end{subfigure}
    \hfill
    \begin{subfigure}{0.49\textwidth}
        \centering
        \includegraphics[width=\textwidth]{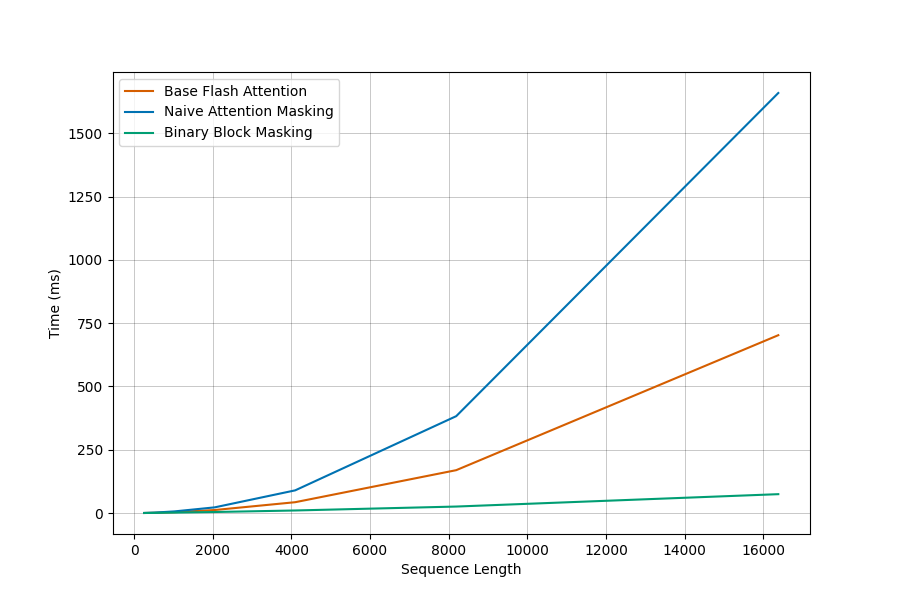}
        \caption{Backward Pass}
    \end{subfigure}
    
    \caption{ALPACA: Sequential Mask}
    \label{fig:Appendix_Alpaca_Sequential}
\end{figure}

\begin{figure}[H]
    \centering
    \begin{subfigure}[b]{0.49\textwidth}
        \centering
        \includegraphics[width=\textwidth]{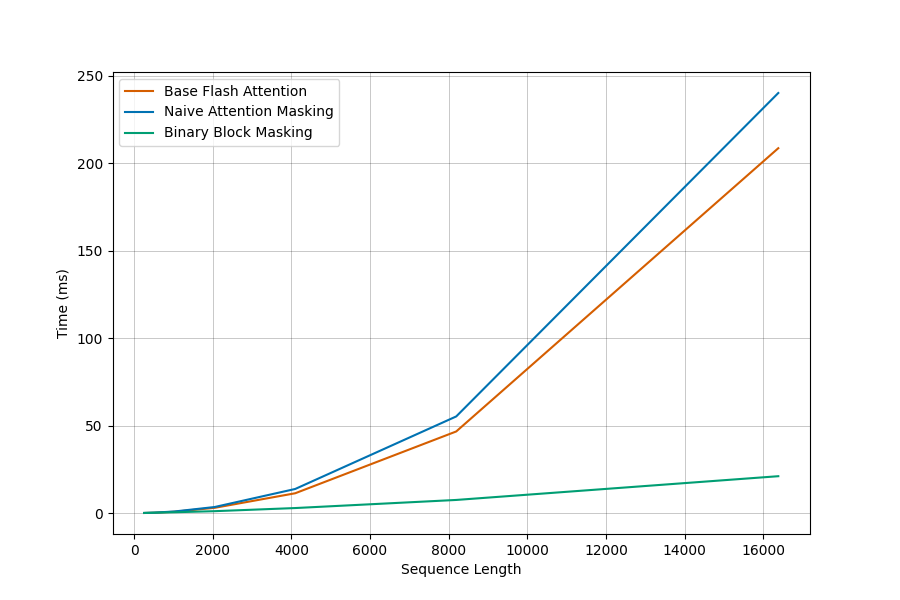}
        \caption{Forward Pass}
    \end{subfigure}
    \hfill
    \begin{subfigure}{0.49\textwidth}
        \centering
        \includegraphics[width=\textwidth]{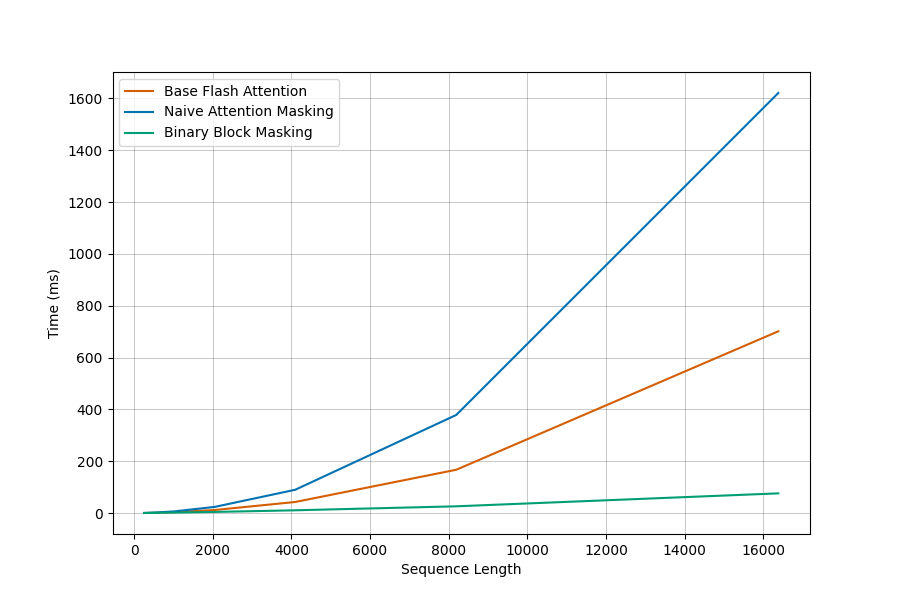}
        \caption{Backward Pass}
    \end{subfigure}
    
    \caption{ALPACA: Input-Bidirectional Mask}
    \label{fig:Appendix_Alpaca_Input-Bidirectional}
\end{figure}

\begin{figure}[H]
    \centering
    \begin{subfigure}[b]{0.49\textwidth}
        \centering
        \includegraphics[width=\textwidth]{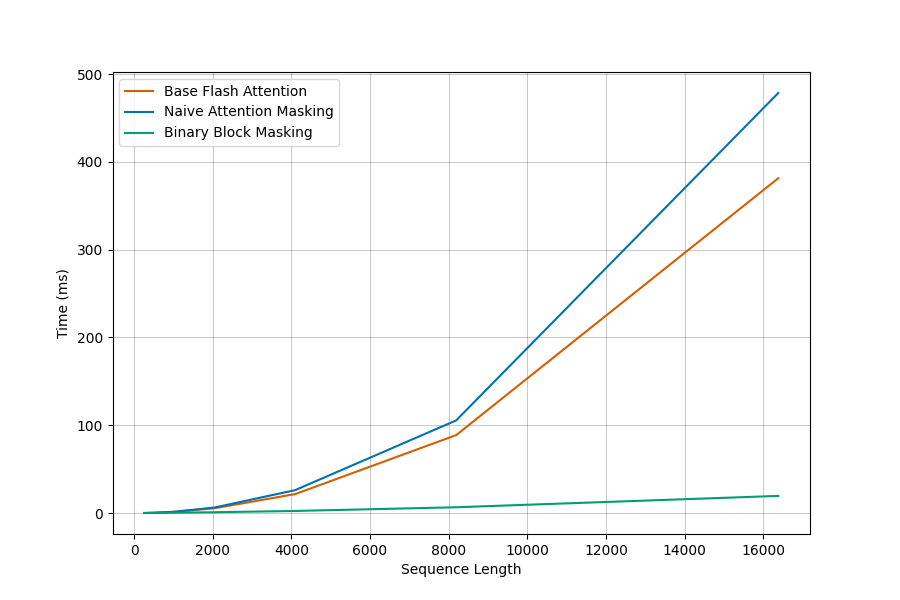}
        \caption{Forward Pass}
    \end{subfigure}
    \hfill
    \begin{subfigure}{0.49\textwidth}
        \centering
        \includegraphics[width=\textwidth]{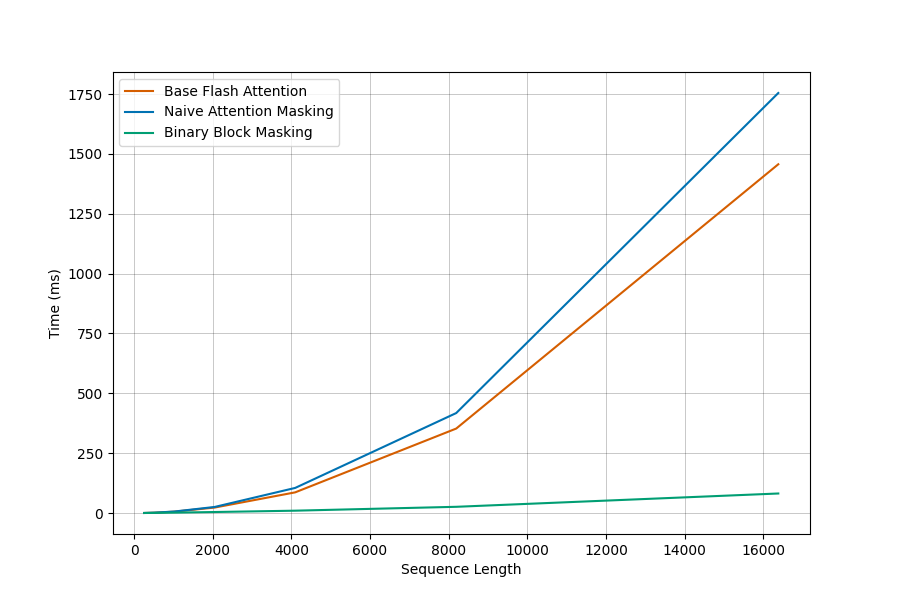}
        \caption{Backward Pass}
    \end{subfigure}
    
    \caption{LongFormer: Windowed Mask}
    \label{fig:Appendix_Windowed}
\end{figure}

\begin{figure}[H]
    \centering
    \begin{subfigure}[b]{0.49\textwidth}
        \centering
        \includegraphics[width=\textwidth]{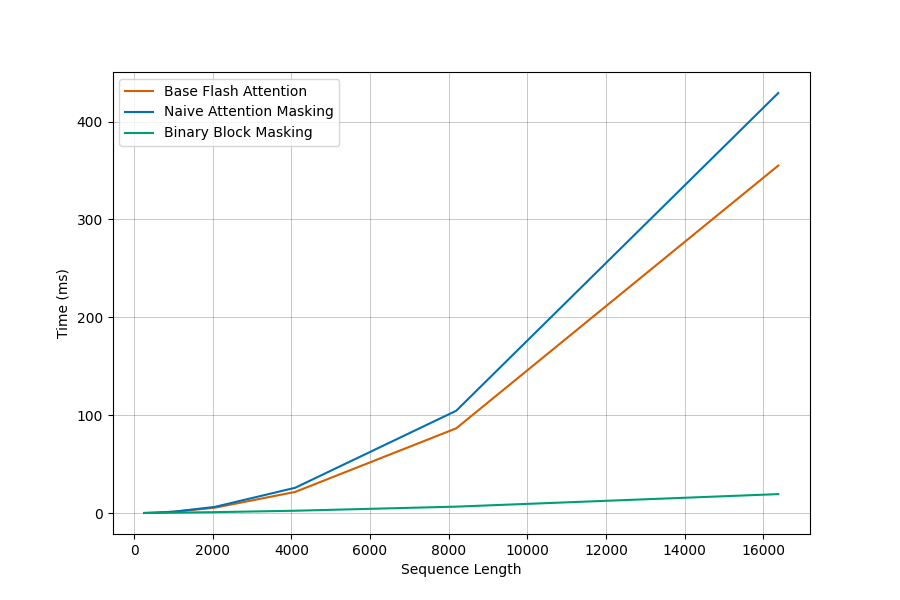}
        \caption{Forward Pass}
    \end{subfigure}
    \hfill
    \begin{subfigure}{0.49\textwidth}
        \centering
        \includegraphics[width=\textwidth]{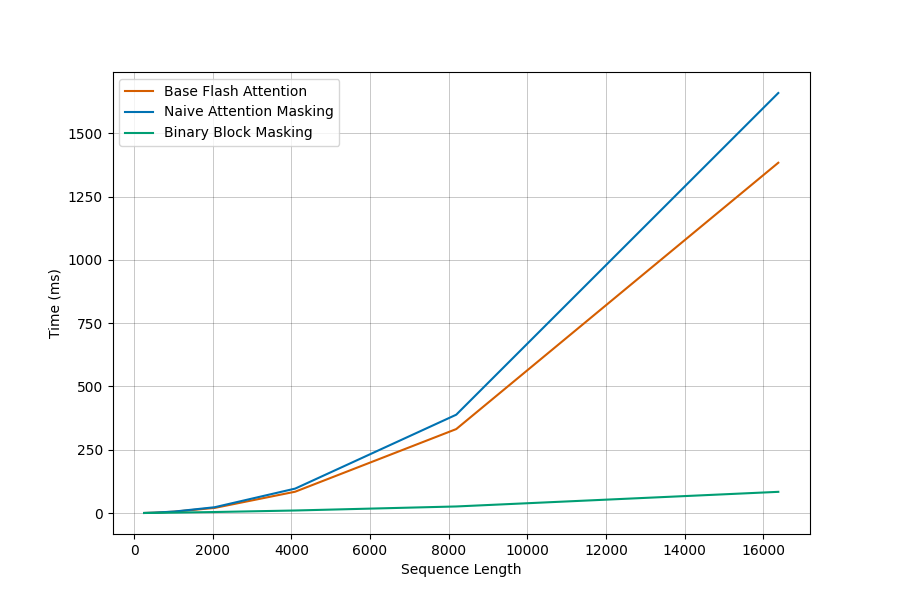}
        \caption{Backward Pass}
    \end{subfigure}
    
    \caption{LongFormer: Dilated Mask}
    \label{fig:Appendix_Dilated}
\end{figure}

\begin{figure}[H]
    \centering
    \begin{subfigure}[b]{0.49\textwidth}
        \centering
        \includegraphics[width=\textwidth]{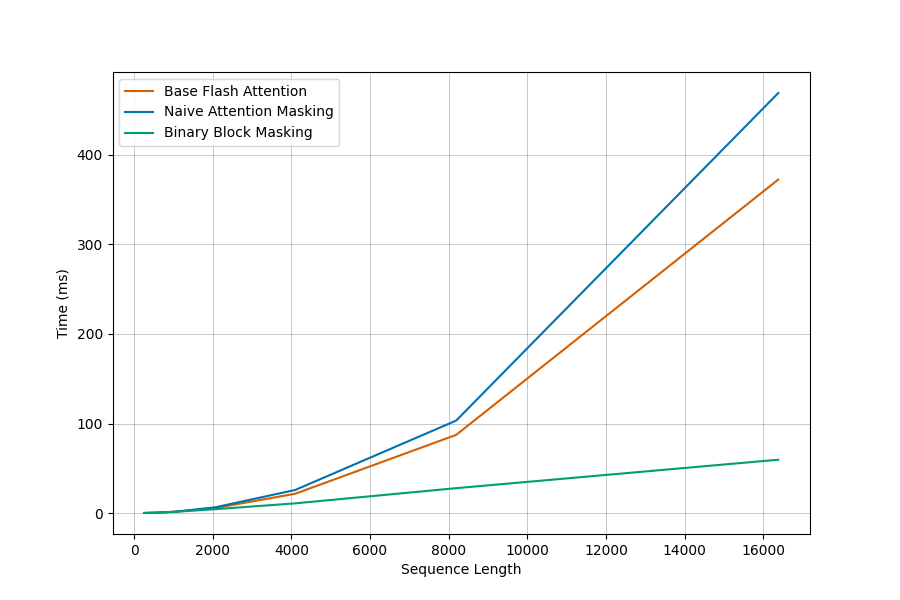}
        \caption{Forward Pass}
    \end{subfigure}
    \hfill
    \begin{subfigure}{0.49\textwidth}
        \centering
        \includegraphics[width=\textwidth]{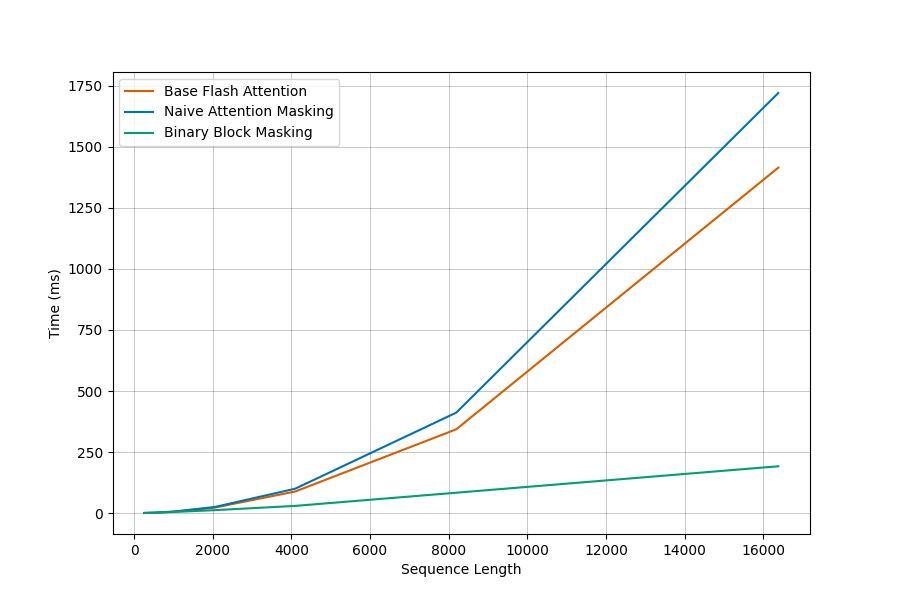}
        \caption{Backward Pass}
    \end{subfigure}
    
    \caption{LongFormer: Global Mask}
    \label{fig:Appendix_Global}
\end{figure}

In the main paper, we reported performance benchmarks based on total runtimes. In this section, we provide a detailed breakdown of the forward and backward pass runtimes for MEDUSA masks\cite{cai2024medusa} (Figures \ref{fig:Appendix_Medusa_Heads} and \ref{fig:Appendix_Medusa_Predicitions}), ALPACA masks\cite{alpaca} (Figures \ref{fig:Appendix_Alpaca_Sequential} and \ref{fig:Appendix_Alpaca_Input-Bidirectional}), and Longformer masks\cite{Beltagy2020Longformer} (Figures \ref{fig:Appendix_Windowed}, \ref{fig:Appendix_Dilated}, and \ref{fig:Appendix_Global})

\section{Dataset and Code License Declaration}
\label{sec:Data_license}
This section provides details on the licenses and sources of the code and datasets utilized in this study.

\begin{enumerate}
\item \textbf{ALPACA}: We obtain the ALPACA dataset from Hugging Face (URL: \url{https://huggingface.co/datasets/tatsu-lab/alpaca}) \cite{alpaca}. The dataset is licensed under the Creative Commons NonCommercial (CC BY-NC 4.0) License.
\item \textbf{MEDUSA}: We have developed our own code to generate MEDUSA masks based on the guidelines provided in the official report \cite{cai2024medusa}.
\item \textbf{Longformer}: We implement our own code for generating Longformer masks, as described in the original research paper \cite{Beltagy2020Longformer}.
\item \textbf{Flash Attention}: For our baseline, we use the public implementation of Flash Attention 2, available at URL: \url{https://github.com/triton-lang/triton/blob/main/python/tutorials/06-fused-attention.py}.
\end{enumerate}

All other code used in this paper was developed by us.

\end{document}